\documentclass[10pt,twocolumn,letterpaper]{article}

\usepackage{wacv}              

\usepackage{graphicx}
\usepackage{amsmath}
\usepackage{amssymb}
\usepackage{booktabs}

\usepackage{enumitem}

\usepackage[pagebackref,breaklinks,colorlinks]{hyperref}

\usepackage{multirow}
\usepackage{float}
\usepackage{booktabs}
\usepackage{amssymb}
\usepackage{pdfrender}
\usepackage{xcolor}
\usepackage{amsmath}
\usepackage{comment}
\usepackage{subcaption} 
\usepackage[accsupp]{axessibility} 

\usepackage[capitalize]{cleveref}
\crefname{section}{Sec.}{Secs.}
\Crefname{section}{Section}{Sections}
\Crefname{table}{Table}{Tables}
\crefname{table}{Tab.}{Tabs.}

\newcommand*{\boldcheckmark}{%
  \textpdfrender{
    TextRenderingMode=FillStroke,
    LineWidth=1.0pt, 
  }{\checkmark}%
}

\def\wacvPaperID{1684} 
\def\confName{WACV}
\def\confYear{2025}

\begin{document}

\title{SADA: Semantic adversarial unsupervised domain adaptation \\ for Temporal Action Localization}

\author{
David Pujol-Perich, Albert Clapés, Sergio Escalera \\
Universitat de Barcelona and Computer Vision Center, Barcelona, Spain\\
{\tt\small \{david.pujolperich, aclapes, sescalera\}@ub.edu}
}
\maketitle

\begin{abstract}
Temporal Action Localization (TAL) is a complex task that poses relevant challenges, particularly when attempting to generalize on new -- unseen -- domains in real-world applications. These scenarios, despite realistic, are often neglected in the literature, exposing these solutions to important performance degradation. In this work, we tackle this issue by introducing, for the first time, an approach for Unsupervised Domain Adaptation (UDA) in sparse TAL, which we refer to as Semantic Adversarial unsupervised Domain Adaptation (SADA). Our contributions are threefold: (1) we pioneer the development of a domain adaptation model that operates on realistic sparse action detection benchmarks; (2) we tackle the limitations of global-distribution alignment techniques by introducing a novel adversarial loss that is sensitive to local class distributions, ensuring finer-grained adaptation; and (3) we present a novel set of benchmarks based on EpicKitchens100 and CharadesEgo, that evaluate multiple domain shifts in a comprehensive manner. Our experiments indicate that SADA improves the adaptation across domains when compared to fully supervised state-of-the-art and alternative UDA methods, attaining a performance boost of up to $6.14\%$ mAP. The code is publicly available at \url{https://github.com/davidpujol/SADA}.
\vspace{-0.15cm}
\end{abstract}
\vspace{-0.3cm}
\section{Introduction}\label{sec:introduction}

\begin{figure}[t]
\centering
\includegraphics[width=0.43\textwidth]{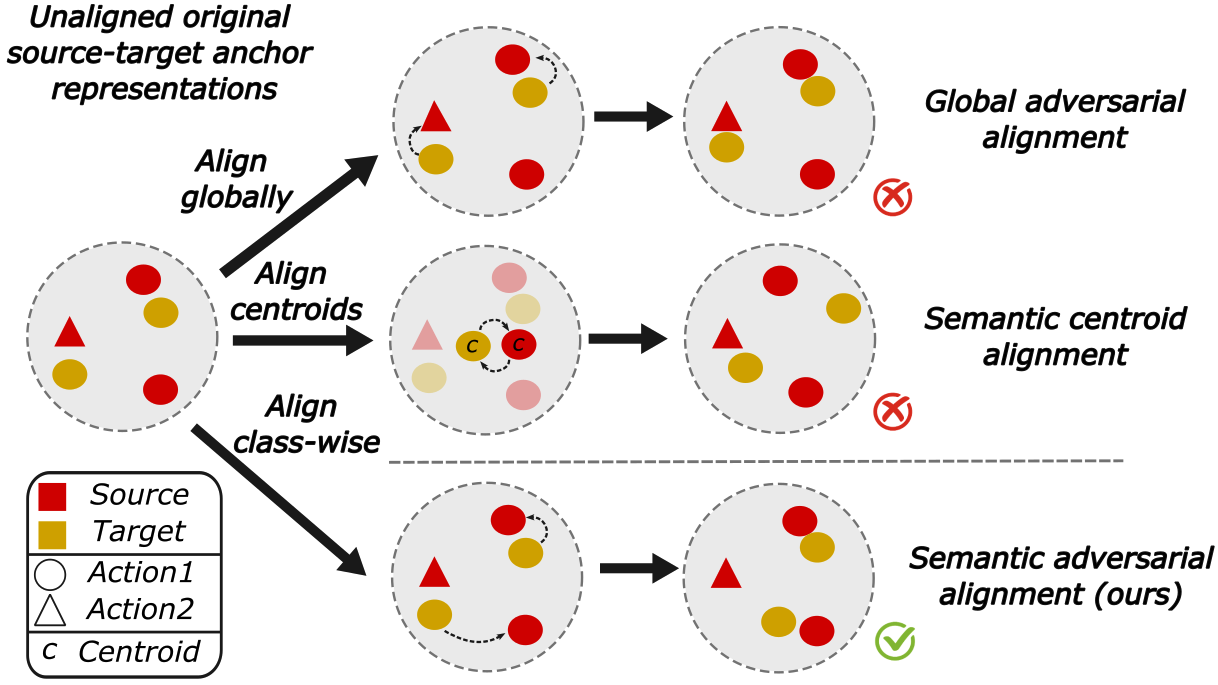}
\vspace{-0.2cm}
\caption{\label{fig:related_work} Illustration of the differences between the two most similar domain-adaptation methods~\cite{ganin2016domain, xie2018learning}, and our proposal, \textit{SADA}. For this, we present a simple scenario with various anchor embeddings of different actions (identified by shapes) and domains (identified by colors). In this scenario, \cite{ganin2016domain} (upper row) aligns embeddings in a class-agnostic manner, making it liable to aligning domain embeddings of unmatching action labels. \cite{xie2018learning} (middle row) computes class-wise mean centroids, and aligns them across domains, but as shown, minimizing their distance does not yield a proper adaptation. \textit{SADA} (last row) improves \cite{ganin2016domain} by aligning class-wise distributions, yielding the correct alignment by not aligning unmatching anchors.}
\label{fig:main_illustration}
\vspace{-0.5cm}
\end{figure}

Recent advances in the field of video understanding have played a critical role in the surge of novel video-based applications -- e.g., video indexing, summarization or recommendation. A critical task of this field is \textit{Temporal Action Localization} (TAL), which involves identifying actions in a video consisting of both their time intervals and action categories. This is particularly difficult given the inherent variabilities of videos. These can be presented, among others, in the form of \textit{appearance variability} -- e.g., different kitchens and/or lighting conditions --, \textit{acquisition variability} -- e.g., different recording devices -- or \textit{viewpoint variability} -- e.g., first- or third-person. All these prompt a certain degree of confusion between similar actions to be discriminated.


Traditionally, fully supervised methods attempt to address this issue by leveraging enough training data to cover all the possible sources of variability. Unfortunately, this becomes virtually impossible when dealing with realistic scenarios. This compels these methods to operate under the influence of unseen data variations -- i.e., \textit{domain gaps} --, which exposes them to a considerable decline in performance. Overcoming this typically involves relabeling data from the new domain, so as to retrain and adapt the model.  Unfortunately, this approach is impractical due to the considerable time and resource consumption involved, a challenge exacerbated when dealing with high-dimensional inputs like videos.

Unsupervised Domain Adaptation (UDA) has recently become a hot topic given its potential to leverage unlabelled data to mitigate this domain-induced degradation \cite{ganin2015unsupervised, motiian2017unified, hassanpour2023survey}. Despite its considerable success in image-based tasks~\cite{csurka2017domain} the application of UDA to video understanding remains underexplored. In fact, to the best of our knowledge, no prior work addresses UDA for TAL setups. The closest proposal is SSTDA~\cite{chen2020action}, which approaches the problem of action segmentation. This focuses on making per-frame action predictions (i.e., \textit{dense TAL}), tackling datasets where no concurrent actions take place~\cite{fathi2011learning,Kuehne12}. This approach inevitably requires additional smoothing techniques to preserve temporal coherence. This motivates the focus of this paper on the more general \textit{sparse TAL} problem --i.e., segment-level predictions -- avoiding the need for additional losses, while intrinsically adapting to multi-label scenarios.

Consequently, in this paper, we propose the first UDA method for sparse multi-label detection on TAL, which we name \textit{Semantic Adversarial unsupervised Domain Adaptation}, or \textit{SADA} for short. Concretely, our proposal specifically builds upon an anchor-based architecture given their recent success on sparse TAL~\cite{zhang2022actionformer, shi2023tridet}. Thus, our goal is to minimize the discrepancy between anchor representations of a labeled source domain and an unlabelled target domain. These anchors are extracted with a multi-resolution architecture~\cite{shi2023tridet} that we couple with a novel adversarial loss that improves the limitations of existing UDA works. Concretely, existing works normally align domain distributions globally~\cite{ganin2015unsupervised}, applying adversarial methods on the feature embeddings regardless of the action class they represent. As we will show, the coarse alignment of anchors of different action classes with even \textit{background} (no-action) anchors can hurt performance. We propose instead to first use pseudo-labeling~\cite{li2023pseudo} to assign an action or background class to each anchor representation. With this, we can factorize the global alignment loss~\cite{ganin2015unsupervised} into independent per-class and background distribution alignments. This results in a more sensitive alignment strategy, less prone to \textit{semantic feature misalignment} -- i.e., semantically meaningless alignments across domains -- and as we will show, better performing.

 Assessing the effectiveness of UDA methods for video understanding is a challenging, still unresolved task. Existing proposals on action segmentation~\cite{chen2020action} follow a subject-based strategy where they aim to adapt a model to new unseen subjects. Here we refer to \textit{subject} as a person appearing in a video. Nevertheless, little data is normally available from a single subject, which inevitably requires grouping several of them for training. This allows the model to generalize over the subject variability under study, making it unsuitable for domain adaptation. To address these limitations, we draw inspiration from the work of \cite{zhang2022audio} on action recognition and investigate the impact of \textit{viewpoint domain shifts} on sparse TAL in CharadesEgo~\cite{sigurdsson2018charades}. However, we contend that a more comprehensive evaluation necessitates setups with more controllable shifts. For this, we also propose a suite of 6 new setups based on EpicKitchens100~\cite{Damen2022RESCALING} which study the effect \textit{appearance} and \textit{acquisition} domain shifts. These benchmarks demonstrate that \textit{SADA} mitigates the performance degradation, improving by a large margin the existing fully supervised (namely \textit{source-only}) and UDA-based proposals. In short, our main contributions are:
\begin{enumerate}[nosep]
    \item We propose for the first time an UDA method suitable for sparse detection scenarios on TAL.
    \item We introduce a novel adversarial loss that factorizes standard global alignment into independent class- and background-wise alignments (see Fig. \ref{fig:main_illustration}).
    \item We present new benchmarks to test sparse detection scenarios when facing 7 different domain shifts, improving the state-of-the-art in all of them.
\end{enumerate}
\section{Related work}\label{sec:related_work}
\vspace{-0.1cm}

\textbf{Temporal Action Localization.} At the time of this writing, most of the literature on the task of \textit{Temporal Action Localization} follows a traditional \textit{source-only} approach. In other words, they restrict the models' visibility solely to a training domain, while other domains seen during testing are not available. These works can be categorized as follows: \textbf{(1) Anchor-based methods}~\cite{buch2017sst, chao2018rethinking,li2021three, lin2021learning, qing2021temporal, sridhar2021class, zhang2022actionformer, shi2023tridet} propose a two-stage pipeline, consisting of a proposal generation and classification. The first applies heuristic methods -- e.g., uniform sampling~\cite{buch2017sst,  zhang2022actionformer} or action boundaries' grouping~\cite{zhao2017temporal, zhao2020bottom} -- to generate a dense set of proposals --i.e., temporal segments. In the second stage, they leverage a learnable classifier to predict the corresponding action class and localization offsets of every anchor. Our work falls into this category motivated by the recent success of these methods achieving state-of-the-art results in many TAL benchmarks~\cite{idrees2017thumos, zhao2019hacs, Damen2022RESCALING}. \textbf{(2) Anchor-free methods}~\cite{shou2017cdc, lin2017single, lin2021learning, yuan2017temporal} avoid this two-stage approach making per-frame predictions of their corresponding action labels. These methods, however, often suffer from a tendency towards over-segmentation given the potential discrepancy between neighboring frames. Consequently, they require often complex smoothing techniques to improve the boundary predictions~\cite{chen2020action}. \textbf{(3) Query-based methods}~\cite{tan2021relaxed, liu2021end, liu2022empirical} recently emerged as an alternative paradigm that follows the principles presented by \cite{carion2020end}. This approach exploits the use of a Transformer encoder-decoder architecture~\cite{vaswani2017attention} to learn a fixed small set of queries given refined video features, each identifying one potential action segment. Intuitively, this results in a non-heuristic-based proposal generation. This comes with the limitation of an increased rigidity, as the number of proposals needs to be fixed beforehand. 

\noindent\textbf{Unsupervised Domain Adaptation.} Domain Adaptation techniques emerge as an effective solution to bridge the gap between data collected from a source and a target distribution, respectively. A large suite of approaches has been proposed to perform this alignment between labeled and unlabeled domains -- e.g., discrepancy minimization~\cite{koniusz2017domain, xu2022aligning} or entropy minimization~\cite{grandvalet2004semi, xu2021partial}. Arguably, nowadays the most popular approach is based on adversarial training~\cite{ganin2015unsupervised,  tzeng2017adversarial, gao2021gradient, huang2023discriminative, hassanpour2023survey}. These learn domain-invariant embeddings~\cite{cui2020gradually} by training in a min-max fashion a domain classifier to discern if samples come from the source or the target domain. Despite convenient, the simplicity of these methods often degrade the quality of the alignment~\cite{li2023pseudo}, as they potentially align embeddings of source and target domain that represent different semantic information -- e.g., different class labels. Few works have been proposed to do this alignment in a more sensitive way~\cite{li2023pseudo}. Works like \cite{xie2018learning, huang2023discriminative}, for instance, reduce the distance of per-class centroids, normally computed as the mean feature embeddings of a given class. Its effectiveness, however, relies on the assumption that the data is distributed somewhat homogeneously around the center, as otherwise, the centroids are not necessarily meaningful. In our work, we couple the advantages of both adversarial domain adaptation and semantic alignment and propose for the first time a pure adversarial semantic loss that yields domain invariant representations in a semantically meaningful way, without making explicit assumptions of the distributions (see Fig.~\ref{fig:related_work}).

\noindent\textbf{Unsupervised Domain Adaptation for TAL.} Despite the considerable success of UDA methods, their applicability has been mostly restricted to image-based scenarios such as image classification~\cite{haeusser2017associative, ganin2015unsupervised, long2015learning, motiian2017unified} or object detection~\cite{chen2018domain, oza2023unsupervised}. Much less attention has been dedicated to video-based applications such as action recognition~\cite{chen2019temporal, jamal2018deep, pan2020adversarial} or spatio-temporal action segmentation~\cite{agarwal2020unsupervised, lu2023exploiting}. To the best of our knowledge, at the time of this writing, there is no direct comparison with our work focusing on UDA for \textit{sparse} TAL. The closest work is SSTDA~\cite{chen2020action} that applies UDA for Action Segmentation. SSTDA proposes the use of two global-distribution-based auxiliary tasks to jointly align cross-domain feature spaces. Unlike our proposal, their work falls into the category of anchor-free, making per-frame action predictions. This restricts its applicability to action segmentation scenarios, where current datasets~\cite{fathi2011learning,Kuehne12} are designed to deal with frame-based single-action classification. In our work, we overcome this limitation by leveraging an anchor-based architecture that enables a natural adaptation to more realistic multi-label scenarios.
\vspace{-0.3cm}
\section{Method}\label{sec:method}
\vspace{-0.1cm}
\begin{figure*}[t]
\centering
\includegraphics[width=0.82\textwidth]{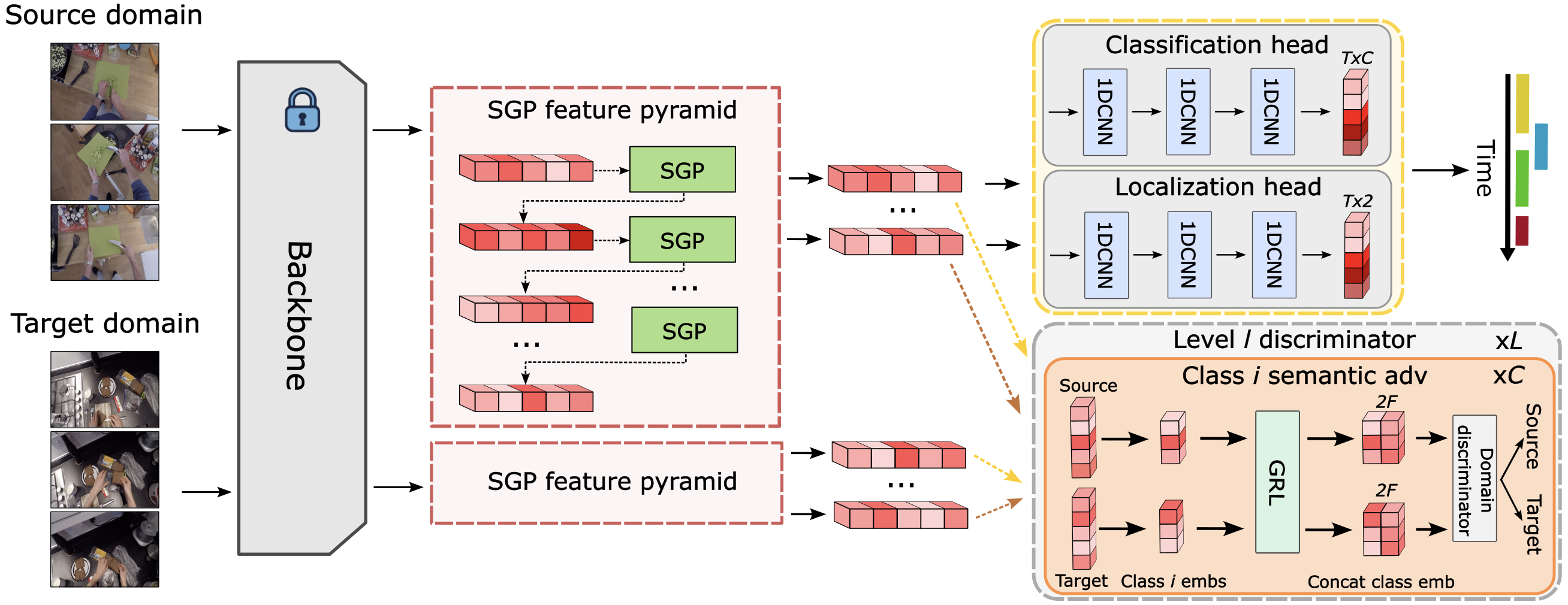}
\vspace{-0.3cm}
\caption{\label{fig:main_architecture} Overview of the main model architecture of SADA. This takes as input videos from a Source and a Target domain, which are both fed to a shared multi-resolution feature extractor pyramid. The output embeddings of both of these domains are then aligned using the semantic alignment loss, \textit{SADA}. This is done with a level and class-wise domain discriminator of the filtered embeddings, based on GT information and pseudo labels, for the source and target domains, respectively. Finally, the resulting domain invariant representations of the source domain are used to train a classification and localization head to learn the underlying task.}
\vspace{-0.4cm}
\end{figure*}

\subsection{Problem definition and notation}\label{sec:notation}
In this paper, we address the problem of unsupervised domain adaptation for TAL. For this, we define a source domain $\mathcal{S}$ and a target domain $\mathcal{T}$. Domain $\mathcal{S}$ consists of $N_\mathcal{S}$ labeled input videos $\{(V_k^\mathcal{S},Y_k^\mathcal{S})\}_{k=1}^{N_\mathcal{S}}$, where each video $V_k^\mathcal{S}$ is a sequence of $T$ frames $(X_{k,1},\dots,X_{k,T})$ with $X_{k,t}\in \mathbb{R}^{H\times W \times C}$. Here $Y_k = \{(b_{k,i}, e_{k,i}, c_{k,i})\}_{i=1}^{G_k}$ contains the begin, end, and class actions of all the ground-truth (GT) segments $G_k$ of the k-th video, respectively. The target domain $\mathcal{T}$ is similar to $\mathcal{S}$ but lacks the GT information. Concretely, it consists of $N_\mathcal{T}$ unlabeled input videos $\{V_k\}_{k=1}^{N_\mathcal{T}}$. Our goal is to train a model that can identify the action segments, including both segment coordinates and action labels, in videos from domain $\mathcal{S}$, while minimizing the performance degradation on the unlabeled domain $\mathcal{T}$.

\subsection{Framework overview}
We propose a model based on a feature pyramid and an anchor-based classification and localization head (see Fig.~\ref{fig:main_architecture}). This architecture is coupled with a novel \textit{semantic adversarial loss} that aligns the anchor embeddings across domains $\mathcal{S}$ and $\mathcal{T}$ in a semantically meaningful way. More in detail, the model takes as input two videos from domain $\mathcal{S}$ and $\mathcal{T}$, respectively. The model first processes the raw input videos using a frozen pre-trained video backbone. The resulting embeddings of both domains are then passed through a shared SGP pyramid~\cite{shi2023tridet} that outputs a set of multi-resolution anchor embeddings for each of the predefined resolution levels. The main goal of our model is to make these anchor embeddings domain invariant. For this, we introduce a level-wise \textit{semantic adversarial loss} that learns in an adversarial manner to align the embeddings of both domains belonging to a given action class $i$ at a given resolution level $l$. Recall that GT information is only available for domain $\mathcal{S}$, therefore we rely on the use of pseudo labeling techniques~\cite{li2023pseudo} to infer the \textit{probable} class labels of the data from domain $\mathcal{T}$. Finally, we use the domain invariant anchors of domain $\mathcal{S}$ to train a classification and localization head that learns the underlying tasks in a standard supervised fashion. In short, this permits to learn a classification and localization head that minimizes the decline of performance when applied to the unseen domain $\mathcal{T}$.

\subsection{Backbone and SGP pyramid}\label{sec:method_backbone}
Our model first takes two input videos $V^\mathcal{S}$ and $V^\mathcal{T}$ of both domains $\mathcal{S}$ and $\mathcal{T}$. For simplicity, both videos $V^\mathcal{S}$ and $V^\mathcal{T}$ have length $T$, which we enforce using padding. The method then processes the two videos applying a frozen pre-trained backbone -- e.g., I3D~\cite{carreira2017quo} or Slowfast~\cite{feichtenhofer2019slowfast}. This permits to extract, in an effective way, temporal cues of the video into a set of refined video features. These embeddings are then fed to an SGP feature pyramid~\cite{shi2023tridet} which combines the use of SGP blocks and the progressive downsampling of the temporal length by a ratio of $2$. This outputs a set of multi-resolution anchor embeddings $Z^\mathcal{S} = \{Z_l^\mathcal{S}\}_{l \in L}$ and $Z^\mathcal{T} = \{Z_l^\mathcal{T}\}_{l \in L}$, for the two domains, respectively. Here $L$ denotes the set of predefined resolution levels and $Z_l \in \mathbb{R}^{T_l \times F}$ the anchor embeddings of level $l$ of a given domain. Concretely, this permits to obtain embeddings for a set of uniformly sampled anchors at each of the $l\in L$ resolution levels. The use of a multi-resolution model is favorable for naturally adapting to different action lengths and abstraction levels.

\subsection{Classification and localization head}\label{sec:method_classification_and_localization}
To learn the underlying TAL task, we train in a fully supervised manner a classification and localization module with the labeled source domain $\mathcal{S}$. Due to the anchor-based nature of our model, we first require a matching strategy between the set of candidate anchors to the actual GT segments. For this, we follow a center sampling strategy \cite{tian2019fcos, shi2023tridet}. In other words, for a given level $l$, we define an anchor as \textit{action anchor} if the time instant $t$ that it represents is near the center of an action. All the rest are marked as \textit{background anchors}. We define $\mathcal{B}_l, \mathcal{E}_l$ and $\mathcal{C}_l$ as the begins, ends and action classes of their matching GT segments. We identify \textit{background anchors} with action label 0. With this, we design a classification head $H_{cls}: \mathbb{R}^{T_l \times F} \rightarrow \mathbb{R}^{T_l \times C}$ that maps each of the anchor embeddings to their class distribution. More specifically, we model this as a sequence of 1D convolutions, and train it using a sigmoid focal loss\cite{lin2017focal}:
\vspace{-0.5cm}
\begin{gather}
\small
\begin{aligned}
\mathcal{L}_{SFL}^l = SFL(H_{cls}(Z^\mathcal{S}_l),\; \mathcal{C}_l).
\end{aligned}
\end{gather} 
Similarly, we model a localization head $H_{loc}: \mathbb{R}^{T_l \times F} \rightarrow \mathbb{R}^{T_l \times 2}$ identically as $H_{cls}$, which predicts the begin-end offsets. We thus define the localization loss as a standard mean squared error (MSE) loss over the \textit{action anchors} only:
\vspace{-0.5cm}
\begin{gather}
\small
\begin{aligned}
\mathcal{L}_{loc}^l = MSE(H_{loc}(Z_{l{_+}}^\mathcal{S}),\; (\mathcal{B}_{l_+} \,||\, \mathcal{E}_{l_+})),
\end{aligned}
\end{gather}
where $l_+$ refers to the filtered \textit{action-anchor} representations from the GT of the $l$-th level only, and $||$ to the concatenation operation. This yields the final task loss defined as:
\begin{gather}
    \small
    \mathcal{L}_{task} = \lambda_{cls} \;\sum_{l \in L} \mathcal{L}_{SFL}^l + \lambda_{loc} \;\sum_{l \in L} \mathcal{L}_{loc}^l.
\end{gather}
Here $\lambda_{cls}$ and $\lambda_{loc}$ are two tunable hyperparameters.

\subsection{Our proposal: Semantic adversarial multi-resolution alignment}\label{sec:method_our_proposal}
\vspace{-0.1cm}
One of the main contributions of this paper is the design of a novel adversarial-based loss that we name \textit{SADA} loss, which attempts to overcome the limitations of the extensively used \textit{global adversarial loss}~\cite{ganin2015unsupervised}. Traditional adversarial domain adaptation relies on the idea of designing a domain classifier that learns to identify the domain that each of the embeddings belongs to. The rest of the model learns concurrently the opposite objective which results in the learning of domain invariant representations\cite{ganin2015unsupervised}. While this approach has been shown to be effective in other fields -- e.g., image classification or object detection -- we find that its performance in more challenging video understanding setups like TAL presents important challenges. One of the main issues, as argued by \cite{li2023pseudo}, is that this loss often suffers from \textit{feature misalignment} which greatly declines its effectiveness. This refers to the cases where these methods align embeddings of non-matching class labels -- i.e., aligning embeddings of domain $\mathcal{S}$ of an action $i$ with embeddings of domain $\mathcal{T}$ of a class $j$. This issue is further exacerbated in TAL given the noise induced by the alignment of the many \textit{background anchors} with the \textit{action anchors}. We study this phenomenon in more depth in the Supp.

\noindent\textbf{Local adversarial alignment:} To fix this \textit{feature misalignment} in realistic scenarios like TAL, we propose an alternative adversarial loss formulation that provides a finer-grained alignment. This loss first attempts to perform a local class-aware alignment. This is, for every given resolution level $l$, we group the \textit{action anchors} -- those matched with a GT action -- of $\mathcal{S}$ and $\mathcal{T}$ according to their action label $i$. This is straightforward for domain $\mathcal{S}$ as we have the GT information. In contrast, for domain $\mathcal{T}$, we use a hard-pseudo labeling strategy~\cite{li2023pseudo} that classifies a given embedding as class $i$ if this is the highest-confidence score of the predicted class distribution, and this is above a threshold $\alpha$. Formally we define the pseudo-label of an anchor $z$ as:
\vspace{-0.5cm}
\begin{gather}
    \small
    \hat{c}_z = {
    \begin{cases}
    \underset{i}{\mathrm{argmax}} \;P_l[z,i] & \text{if } P_l[z,i] > \alpha \\
    0 & \text{otherwise},
    \end{cases}
    }
    \label{eq:pseudo_labeling}
\end{gather}
where $P_l = H_{cls}(Z^\mathcal{T}_l) \in \mathbb{R}^{T_l \times C}$ are the predicted class probabilities of the anchors. Notice that we mark with class 0 the \textit{background anchors}, which are not assigned to any action class. From this, we obtain the newly grouped embeddings of source and target domain of class \textit{i} on level \textit{l}
\vspace{-0.5cm}
\begin{gather}
    \small
    A_i^l = \{Z_l^{\mathcal{S}}[z] \;:\; c_z = i \}_{z \in T_l}, \\
    \small
    B_i^l = \{Z_l^{\mathcal{T}}[z] \;:\; \hat{c}_z = i \}_{z \in T_l},
\end{gather}
for $A_i^l \in \mathbb{R}^{T_{l,i} \times F}$, $B_i^l \in \mathbb{R}^{T_{l,i} \times F}$. Also, $c_z$ is the GT action label of anchor $z$ and $\hat{c}_z$ is its computed pseudo-label from Eq.~\ref{eq:pseudo_labeling}. We then adversarially train a single domain classifier $D: \mathbb{R}^{2F} \rightarrow \{0,1\}$ to identify the domain of each of these embeddings using a binary cross entropy (BCE) loss:
\vspace{-0.2cm}
\begin{gather}
\small
\begin{aligned}
\mathcal{L}_{local}^l = \sum_{i=1}^C (\mathcal{L}_{\text{\tiny{BCE}}}(D(A_i^l\, || \,E_i), d_\mathcal{S})) + (\mathcal{L}_{\text{\tiny{BCE}}}(D(B_i^l\, ||\, E_i), d_\mathcal{T}))
\end{aligned}
\label{eq:local_alignment},
\end{gather}
where $d_\mathcal{S}$ and $d_\mathcal{T}$ are the domain labels.
We then introduce a Reverse Gradient Layer (GRL)\cite{ganin2015unsupervised} before the discriminator $D$ to invert the gradients sign, creating a min-max game where the feature extractor learns to \textit{confuse} the discriminator. We condition the discriminator to class $i$ using a learnable class embedding $e_i \in \mathbb{R}^{F}$ that we replicate for every selected anchor into an embedding $E_i$ (see Supp. for an ablation of this model decision).

\noindent\textbf{Local and global alignment (\textit{SADA}):} Eq.~\ref{eq:local_alignment} aims solely to align the \textit{action anchors} -- which are classified as one of the $C$ classes -- but \textit{what happens with the \textit{background} embeddings that fall below the threshold $\alpha$?} In this case, the loss ignores their influence, yielding only partial alignment.

To overcome this issue, we propose our final \textit{SADA} loss which attempts to combine the best of both \textit{global alignment loss} \cite{ganin2015unsupervised} and Eq. \ref{eq:local_alignment}. For this, we introduce a new loss term for the \textit{background anchors} as follows:
\vspace{-0.2cm}
\begin{gather}
    \small
    \begin{aligned}
    \mathcal{L}_{bkg}^l = \mathcal{L}_{\text{BCE}}(D(A_0^l\; ||\; E_0), d_\mathcal{S}) + \mathcal{L}_{\text{BCE}}(D(B_0^l\;||\; E_0), d_\mathcal{T}),
    \end{aligned}
    \label{eq:bkg_loss}
\end{gather}
where again $A_0^l$ and $B_0^l$ are the selected \textit{background anchors}, and $E_0\in\mathbb{R}^F$ is the learnable \textit{background} embedding.  Coupling Eq. \ref{eq:local_alignment} and Eq. \ref{eq:bkg_loss} yields the final formulation of our proposed loss, combining \textit{local} (class-wise) alignment with the \textit{background} anchors alignment. Formally:
\vspace{-0.1cm}
\begin{gather}
    \small
    \mathcal{L}_{sada} = \sum_{l \in L} \lambda_l \; (\mathcal{L}_{local}^l + \mathcal{L}_{bkg}^l),
    \label{eq:sada_loss}
\end{gather}
where $\lambda_l$ is a hyper-parameter that modulates the importance of level $l$ on the final alignment loss. See Supp. for an analysis of our model's sensitivity to this parameter choice.

\subsection{Training}
During training, we formulate the final loss as a min-max game where the main model architecture is optimized over the classification and localization loss while maximizing the adversarial loss. In parallel, the discriminator model $D$ attempts to minimize the discriminator loss only. Formally,
\vspace{-0.66cm}
\begin{gather}
    \small
    \mathcal{L} = \lambda_{task}\mathcal{L}_{task} + \lambda_{sada} \mathcal{L}_{sada}.
\end{gather}
Note again $\mathcal{L}_{task}$ is optimized with domain $\mathcal{S}$ while $\mathcal{L}_{sada}$ promotes the alignment between both domains $\mathcal{S}$ and $\mathcal{T}$. Moreover, $\lambda_{task}$ and $\lambda_{sada}$ are tunable parameters.

\section{Datasets and experiments}\label{sec:experimentation}
In this section, we present a novel comprehensive benchmark to evaluate the task of UDA for TAL. Our setup, for the first time, goes beyond action segmentation and evaluates the adaptation to different domain shifts in more realistic scenarios with sparse multi-label annotations. We then showcase the effectiveness of our model over the state-of-the-art methods together with several relevant ablations.

\subsection{Benchmarks for UDA on sparse TAL}\label{sec:experimentation_experimental_setup}

Evaluating domain adaptation-based methods in the context of video understanding is a challenging issue that requires the definition of a reasonable domain gap and identifying a sufficiently large set of intersecting action classes. Dividing existing datasets into different domains that comply with these conditions often restricts the amount of data to learn and adapt. SSTDA~\cite{chen2020action} approaches this problem on GTEA~\cite{fathi2011learning} and Breakfast~\cite{Kuehne12} by defining a subject-based partitioning where they aim to adapt the model to new unseen subjects. However, as little data is available from a single subject in those datasets, they group several users for training. This allows the model to generalize over the subject variability under study, making it unsuitable to test for domain adaptation. Closely related to our work, \cite{wei2023unsupervised, munro2020multi} propose several UDA scenarios for video classification based on EpicKitchens100~\cite{Damen2022RESCALING}. These define 3 domains based on the data of 3 different kitchens, thus performing cross-kitchen evaluation. This limits even further the amount of data in each domain -- i.e., between 15 to 29 videos per domain.

\begin{figure}[t]
\centering
\includegraphics[width=0.4\textwidth]{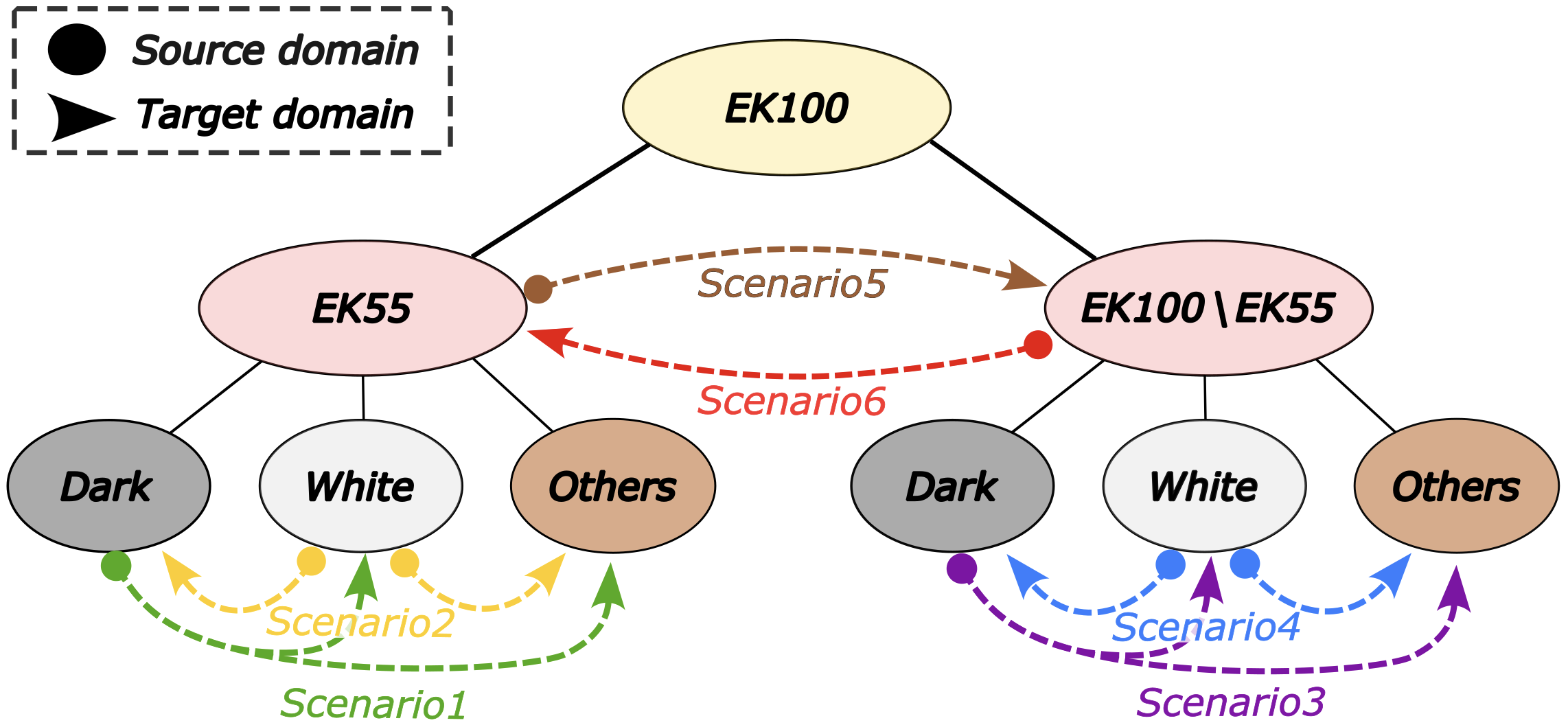}
\vspace{-0.2cm}
\caption{\label{fig:experimental_setup_illustration} Overview of the 6 proposed experimental setups for EpicKitchens100. Concretely, S1 and S2 evaluate the videos from the original EK55. They define the dark-counter and white-counter kitchens as Source, respectively, and the rest as Target. S3 and S4 are similar except that they consider only the \textit{newest} videos from EK100. S5 and S6 use the \textit{old} videos as Source and the \textit{new} videos as Target, and vice versa.} 
\vspace{-0.5cm}
\end{figure}

\noindent\textbf{EpicKitchens100:} To overcome these limitations, we first propose a new set of 6 different scenarios \textit{(S1, \dots, S6)} for sparse TAL based on EK100~\cite{Damen2022RESCALING}, (see Fig.~\ref{fig:experimental_setup_illustration}). EK100 presents an ideal base for our tasks as it has become a gold standard to evaluate complex sparse detection scenarios on long egocentric videos (up to 45 minutes). We identify two domain gaps in this dataset: an \textit{appearance domain shift} based on the different colors of the kitchen counters; and an \textit{acquisition domain shift} that results in the differences of lighting and camera conditions when extending EK55 \cite{Damen2022RESCALING} into its new version EK100 \cite{Damen2022RESCALING}. This permits to define a rich set of benchmarks that provide a more fine-grained and comprehensive evaluation. Importantly, this strategy is also suitable for single-source settings which do not allow an easy generalization over the shift under study. For example, if a model is trained with dark-counter kitchens only, it cannot easily generalize to white-counter kitchens.

More in detail, we introduce 4 different benchmarks that derive from the aforementioned \textit{appearance shift}. For this, we first define two different splits of \textit{EK55} and \textit{\{EK100 \textbackslash EK55\}} videos according to \cite{Damen2022RESCALING}, put more plainly \textit{old} versus \textit{new} videos. We then further split each of them according to the color of the kitchen counter and define 4 different setups of single-source domain and multi-target domain -- i.e., \textit{dark} $\rightarrow$ \textit{white} and \textit{other} kitchens; and \textit{white} $\rightarrow$ \textit{dark} and \textit{other} kitchens, for \textit{old} and \textit{new} videos, respectively.  Similarly, given the two splits between \textit{EK55} and \textit{\{EK100 \textbackslash EK55\}} videos, we propose 2 additional setups of \textit{EK55} $\rightarrow$ \textit{\{EK100 \textbackslash EK55\}} and \textit{\{EK100 \textbackslash EK55\}} $\rightarrow$ \textit{EK55}. These measure the adaptability to different acquisition conditions -- i.e., lighting and camera conditions. This results in domains that contain in the order of hundreds of videos. See Supp. for the domain visualizations and statistics.

One important consideration is that EK100 exhibits a very strong long-tail label distribution, where the vast majority of its 97 actions represent only a marginal percentage of the overall data (see Supp.). Adapting DA methods to these long tail distributions falls beyond the scope of this paper and most of the existing literature. For this reason, we follow a similar approach to \cite{munro2020multi} and limit our evaluation to the 10 majority classes for both domains $\mathcal{S}$ and $\mathcal{T}$, representing $80\%$ of the original data. This still results in a very complex task, proof of which is that the best-performing models to date attain less than a 30\% mAP~\cite{zhang2022actionformer, shi2023tridet}.

\noindent\textbf{CharadesEgo:} One important aspect of the 6 previous setups is that all the domains share an egocentric viewpoint. For this reason, we evaluate the model degradation under extreme domain shifts caused by major changes in the perspective. For this, we use CharadesEgo~\cite{sigurdsson2018charades} which extends the original third-person videos from Charades~\cite{sigurdsson2016hollywood} by matching them with their corresponding egocentric videos. Following \cite{zhang2022audio}, we define the Source domain as the third-person videos, and the Target domain as the egocentric ones. Given the extreme shift that these present, we limit the task to predicting the verb of each of the action segments, and similarly to the previous case, we keep the 10 majority verbs only. We refer to the Supp. for more details.

\subsection{Experimental results}\label{sec:experimental_results}

\begin{table*}[t]
\centering
\resizebox{\textwidth}{!}{%
\footnotesize
    \renewcommand{\arraystretch}{0.9}
\centering
\setlength{\tabcolsep}{4pt} 
\begin{tabular}{c | l | ccccc | c || c | l | ccccc | c}
\toprule
\textbf{Scenario} & \textbf{Model} & \multicolumn{5}{c|}{\textbf{mAP \{10,20,30,40,50\}\%}} & \textbf{Avg} & \textbf{Scenario} & \textbf{Model} & \multicolumn{5}{c|}{\textbf{mAP \{10,20,30,40,50\}\%}} & \textbf{Avg} \\
\midrule
\multirow{6}{*}{S1} & Actionformer~\cite{zhang2022actionformer} & 30.21 & 28.73 & 26.39 & 22.60 & 17.09 & 25.00 & \multirow{6}{*}{S2} & Actionformer~\cite{zhang2022actionformer} & 27.46 & 26.54 & 24.61 & 21.85 & 17.19 & 23.53 \\
& Tridet~\cite{shi2023tridet} & 29.87 & 28.39 & 25.97 & 22.06 & 16.94 & 24.65 & & Tridet~\cite{shi2023tridet} & 30.03 & 28.97 & 26.96 & 23.48 & 18.18 & 25.52 \\
& Ours (src-only) & 30.74 & 29.47 & 27.23 & 23.53 & 18.07 & 25.80 & & Ours (src-only) & 29.65 & 28.69 & 26.86 & 23.88 & 19.14 & 25.64 \\
& Actionformer+SADA & 31.71 & 30.30 & 27.95 & 24.08 & 18.69 & 26.55 & & Actionformer+SADA & 29.84 & 28.89 & 26.83 & 23.76 & 18.85 & 25.63 \\
& Tridet+SADA & 29.55 & 28.27 & 26.16 & 22.89 & 17.63 & 24.90 & & Tridet+SADA & 30.21 & 29.33 & 27.58 & 24.35 & 19.44 & 26.18 \\
& \textbf{Ours (SADA)} & \textbf{31.60} & \textbf{30.29} & \textbf{28.22} & \textbf{24.47} & \textbf{18.98} & \textbf{26.72} & & \textbf{Ours (SADA)} & \textbf{31.54} & \textbf{30.68} & \textbf{28.77} & \textbf{25.52} & \textbf{20.22} & \textbf{27.34} \\
\midrule
\multirow{6}{*}{S3} &
Actionformer~\cite{zhang2022actionformer} & 28.11 & 26.94 & 24.89 & 21.43 & 16.51 & 23.57 & \multirow{6}{*}{S4} & Actionformer~\cite{zhang2022actionformer} & 33.52 & 32.31 & 29.84 & 26.48 & 20.11 & 28.45 \\
& Tridet~\cite{shi2023tridet} & 29.47 & 28.32 & 25.50 & 21.99 & 16.34 & 24.32 & & Tridet~\cite{shi2023tridet} & 34.01 & 32.52 & 30.07 & 26.40 & 19.78 & 28.55 \\
& Ours (src-only) & 30.03 & 28.70 & 26.62 & 23.03 & 17.79 & 25.23 & & Ours (src-only) & 34.41 & 33.38 & 30.58 & 26.99 & 21.00 & 29.27 \\
& Actionformer+SADA & 30.54 & 29.31 & 27.36 & 23.75 & 18.80 & 25.95 & & Actionformer+SADA & 34.11 & 32.87 & 30.60 & 27.05 & 20.93 & 29.11 \\
& Tridet+SADA & 31.34 & 30.16 & 27.94 & 24.48 & 19.31 & 26.64 & & Tridet+SADA & 34.50 & 33.19 & 30.65 & 27.25 & 20.83 & 29.29 \\
& \textbf{Ours (SADA)} & \textbf{32.69} & \textbf{31.49} & \textbf{29.17} & \textbf{25.51} & \textbf{19.72} & \textbf{27.72} & & \textbf{Ours (SADA)} & \textbf{34.86} & \textbf{33.73} & \textbf{31.16} & \textbf{27.45} & \textbf{21.46} & \textbf{29.73} \\
\midrule
\multirow{6}{*}{S5} & Actionformer~\cite{zhang2022actionformer} & 22.87 & 21.87 & 20.10 & 17.23 & 13.33 & 19.08 & \multirow{6}{*}{S6} & Actionformer~\cite{zhang2022actionformer} & 22.16 & 21.22 & 19.71 & 17.44 & 14.08 & 18.92 \\
& Tridet~\cite{shi2023tridet} & 24.77 & 22.93 & 21.49 & 19.09 & 15.15 & 20.48 & & Tridet~\cite{shi2023tridet} & 22.47 & 21.57 & 20.19 & 17.87 & 14.41 & 19.30 \\
& Ours (src-only) & 25.58 & 24.79 & 23.08 & 19.56 & 15.15 & 21.63 & & Ours (src-only) & 20.96 & 20.22 & 19.08 & 16.97 & 14.09 & 18.27 \\
& Actionformer+SADA & 24.85 & 23.99 & 22.21 & 19.11 & 15.28 & 21.09 & & Actionformer+SADA & 23.05 & 22.10 & 20.71 & 18.31 & 14.72 & 19.78 \\
& Tridet+SADA & 24.88 & 24.00 & 22.20 & 19.02 & 14.88 & 21.00 & & Tridet+SADA & 21.00 & 20.08 & 18.64 & 16.39 & 13.13 & 17.85 \\
& \textbf{Ours (SADA)} & \textbf{25.93} & \textbf{25.06} & \textbf{23.47} & \textbf{20.45} & \textbf{16.12} & \textbf{22.21} & & \textbf{Ours (SADA)} & \textbf{23.94} & \textbf{22.95} & \textbf{21.47} & \textbf{19.16} & \textbf{15.24} & \textbf{20.55} \\
\bottomrule
\end{tabular}}
\vspace{-0.2cm}
\caption{Comparison with SOTA for the 4 appearance-shift scenarios (1-4) and the 2 acquisition-shift scenarios (5-6) on EpicKitchens100.}
\vspace{-0.4cm}
\label{tab:appearance_main_results}
\end{table*}

In this section, we present the main experimental results evaluating \textit{appearance} and \textit{acquisition shifts}. All these experiments follow the standard transductive unsupervised DA protocol~\cite{pan2009survey, csurka2017comprehensive}, and report the mean average precision (mAP) at different intersections over union (IOU) thresholds ($10\% - 50\%$).

\noindent\textbf{EpicKitchens100.} In Tab. \ref{tab:appearance_main_results} we first present the results obtained in the 4 different scenarios that we designed to evaluate the performance of our method when facing different \textit{appearance shifts} induced by changes in the background information. Concretely, we first evaluate the performance of the Actionformer~\cite{zhang2022actionformer} and TriDet~\cite{shi2023tridet}, the two best-performing methods on EK100 dataset~\cite{Damen2022RESCALING}, as well as our proposed architecture without our SADA loss. We then compare these 3 architectures with their extensions which include the \textit{SADA} loss. Observe that our proposed loss improves the source-only (SO) version of all the architectures in all these 4 scenarios, showing robustness across the chosen underlying architecture. For instance on S3, our proposed loss yields an improvement of up to $2.49\%$ mAP over its respective SO version. Additionally, we observe that our final architecture -- i.e., Ours(SADA) -- improves the best-performing existing SO method -- i.e., Tridet\cite{shi2023tridet} -- by up to a $3.4\%$ mAP for the \textit{black-counter kitchens} scenarios -- i.e., S1 and S3 -- and $1.82\%$ mAP for the \textit{white-counter kitchens} --i.e., S2 and S4. Similarly, the last two scenarios of Tab. \ref{tab:appearance_main_results} present a similar behavior than before. In all but one case on the Tridet~\cite{shi2023tridet} architecture, the use of the SADA loss yields a performance gain of up to $2.28\%$ mAP. This is a relative $12.48\%$ improvement. Our final model, moreover, improves the best-performing existing baseline -- i.e., Tridet -- by $1.73\%$ mAP and $1.25\%$ mAP, respectively.

\begin{table}[t]
\centering
\resizebox{0.44\textwidth}{!}{%
\footnotesize
    \renewcommand{\arraystretch}{0.9}
\centering
\setlength{\tabcolsep}{4pt} 
\begin{tabular}{l | ccccc | c}
\toprule
\textbf{Model} & \multicolumn{5}{c|}{\textbf{mAP \{10,20,30,40,50\}\%}} & \textbf{Avg} \\
\midrule
Actionformer~\cite{zhang2022actionformer} & 31.22 & 28.51 & 23.82 & 18.91 & 13.94 & 23.28 \\
Tridet~\cite{shi2023tridet} & 30.18 & 27.06 & 22.58 & 17.81 & 13.10 & 22.15 \\
Ours (src-only) & 30.68 & 27.74 & 23.41 & 18.46 & 13.58 & 22.77 \\ 
Actionformer+SADA & 31.46 & 28.57 & \textbf{24.17} & 19.04 & 13.87 & 23.42 \\
Tridet+SADA & 31.53 & 28.41 & 24.04 & 18.88 & 13.98 & 23.37 \\
\textbf{Ours (SADA)} & \textbf{31.68} & \textbf{28.64} & 24.09 & \textbf{19.06} & \textbf{14.14} & \textbf{23.52} \\
\bottomrule
\end{tabular}}
\vspace{-0.2cm}
\caption{Comparison with the state-of-the-art on CharadesEgo.}
\vspace{-0.5cm}
\label{tab:charadesego_results}
\end{table}

\noindent{\textbf{CharadesEgo.} In Tab. \ref{tab:charadesego_results} we report a similar experimental comparison of our proposed loss when evaluated on CharadesEgo. Concretely, we showcase the performance boost that SADA reports on the three test architectures -- i.e., Actionformer~\cite{zhang2022actionformer}, Tridet~\cite{shi2023tridet} and Ours. Observe that SADA, for instance, improves the performance of Tridet~\cite{shi2023tridet} by $1.37\%$ mAP. Similarly, our proposed architecture \textit{Ours (SADA)} improves the SO version by  $0.75\%$ mAP and yields the overall highest scores.
\vspace{-0.1cm}

\subsection{Ablation studies}\label{sec:ablation_studies}
\vspace{-0.1cm}
\noindent\textbf{Comparing to other domain adaptation methods.} In Sec.~\ref{sec:experimental_results} we showed that \textit{SADA} consistently improves the performance of the three tested state-of-the-art SO architectures when evaluated on our newly proposed setups. The question remains however of how well does \textit{SADA} perform compared to existing domain adaptation methods. In this regard, following existing video-based domain adaptation works, we first evaluate various canonical UDA domain adaptation methods --i.e., DANN~\cite{ganin2016domain}, ADDA~\cite{tzeng2017adversarial}, WDGRL~\cite{shen2018wasserstein}, MSTN~\cite{xie2018learning}, FGDA~\cite{gao2021gradient} and DRDA~\cite{huang2023discriminative}. These methods are integrated into our proposed underlying architecture resulting in a fair comparison with \textit{SADA}. Additionally, we find that to the best of our knowledge, there is no existing UDA method for TAL in the literature that is directly comparable to us. Nevertheless, to provide a richer comparison, we adapt the closest action segmentation proposal, SSTDA~\cite{chen2020action} and a state-of-the-art domain-adaptation method for video classification -- i.e., TranSVAE~\cite{wei2023unsupervised} -- to our proposed setup (we refer to the Supp. for all the details of all these baselines). Concretely, in Tab.~\ref{tab:ablation_baselines} we compare \textit{SADA} with these baselines on S3. These results empirically demonstrate the effectiveness of our method by attaining the best results over all tested UDA methods. More in detail, \textit{SADA} yields an improvement of up to $0.7\%$ mAP over the second best-performing method (ADDA). Find the complete ablation for the other scenarios in the Supp.

\begin{table}[t]
\centering
\footnotesize
    \renewcommand{\arraystretch}{0.9}
\resizebox{0.32\textwidth}{!}{%
\begin{tabular}{c | ccc | c}
\toprule
\textbf{Model} & \multicolumn{3}{c|}{\textbf{mAP \{10,30,50\}\%}} & \textbf{Avg} \\
\midrule
DANN \cite{ganin2016domain} & 30.48 & 27.07 & 18.12 & 25.22 \\
ADDA \cite{tzeng2017adversarial} & 31.84 & 28.53 & 19.11 & 26.49 \\
WDGRL \cite{shen2018wasserstein} & 25.05 & 22.08 & 13.91 & 20.35\\
FGDA \cite{gao2021gradient} & 26.21 & 22.99 & 14.12 & 21.10 \\
DRDA \cite{huang2023discriminative} & 30.79 & 26.97 & 18.60 & 25.45 \\
MSTN \cite{xie2018learning} & 31.07 & 27.16 & 17.21 & 25.14\\
SSTDA \cite{chen2020action} & 31.17 & 28.01 & 18.98 & 26.05 \\
TranSVAE \cite{wei2024unsupervised} & 29.91 & 26.12 & 16.16  & 24.06\\
\midrule
\textbf{Ours (SADA)} & \textbf{32.69} & \textbf{29.17} & \textbf{19.72} & \textbf{27.19} \\
\bottomrule
\end{tabular}}
\vspace{-0.2cm}
\caption{Comparison with SOTA UDA methods on S3.}
\vspace{-0.6cm}
\label{tab:ablation_baselines}
\end{table}

\noindent\textbf{Analysis of our loss.}
Next, we ablate over different variants of our proposed \textit{SADA} loss (see Eq. \ref{eq:sada_loss}). Concretely, in Tab. \ref{tab:ablation_loss} we study the effect of class-wise distribution alignment (Eq. \ref{eq:local_alignment}), global distribution alignment (equivalent to DANN \cite{ganin2016domain}) and semantic background alignment (Eq. \ref{eq:bkg_loss}). In this regard, we highlight that the global adaptation seems to consistently improve upon aligning only background embeddings. This is because the latter yields only a partial alignment of the embeddings, not considering \textit{class anchors}. Therefore, a \textit{rougher} yet complete adaptation might seem beneficial. We also observe that aligning local class-wise distributions has a considerable positive effect. Its effect, however, considerably decreases when combined with a global alignment \cite{ganin2016domain}, as all the \textit{class anchors} are then subject to the concurrent alignments of domain-level and class-wise distributions. This observation is also consistent with the performance decrease that we observe when combining global and background alignment, which again suggests that the concurrent alignment of background embeddings with two adaptation losses is harmful to performance. Finally, we observe that we consistently obtain the best results when the local alignment loss is coupled with the complementary (non-overlapping) background loss -- i.e., SADA loss -- indicating that this semantic fine-grained, yet complete, alignment is the most desirable approach. Find in the Supp. the complete ablations for additional scenarios and the class-wise analysis.

\begin{table}[t]
\centering
\footnotesize
    \renewcommand{\arraystretch}{0.9}
\resizebox{0.35\textwidth}{!}{%
\begin{tabular}{ccc|ccc|c}
\toprule
\textbf{Local} & \textbf{Global} & \textbf{Bkg} & \multicolumn{3}{c|}{\textbf{mAP \{10,30,50\}\%}} & \textbf{Avg} \\
\midrule
 & & & 30.03 & 26.62 & 17.79 & 24.81  \\
 & $\boldcheckmark$ & & 30.48 & 27.07 & 18.12 & 25.22  \\
 &  & $\boldcheckmark$ & 30.34 & 26.66 & 17.04 & 24.68\\
 & $\boldcheckmark$ & $\boldcheckmark$ & 29.76 & 26.63 & 17.52 & 24.64\\
$\boldcheckmark$ & & & 31.36 & 28.01 & 18.67 & 26.01 \\
$\boldcheckmark$ & $\boldcheckmark$ & & 30.09 & 26.88 & 17.43 & 24.80  \\
\midrule
$\boldcheckmark$ & & $\boldcheckmark$ & \textbf{32.69} & \textbf{29.17} & \textbf{19.72} & \textbf{27.19} \\
\bottomrule

\end{tabular}}%
\vspace{-0.2cm}
\caption{Ablation of the effect of the components of \textit{SADA} on S3.}
\vspace{-0.2cm}
\label{tab:ablation_loss}
\end{table}

\noindent\textbf{Analysis of the impact of \textit{background anchors.}} One critical aspect of anchor-based methods for TAL is the presence of numerous \textit{background anchors}. We argue that the confusion that these embeddings induce is one of the main challenges for the transfer to a different domain. Concretely, we hypothesize this is because of their high intra-class variance, and their low inter-class variance with other action classes --i.e., they share aspects like appearance. Nevertheless, in this ablation we show that SADA partially mitigates this effect. For this, we design an experiment that artificially masks out all the \textit{background anchors} during inference to elucidate the ideal performance if we could ignore entirely the confusion they generate --i.e., by being wrongly predicted as an action. Concretely, in Tab. \ref{tab:ablation_effect_background_anchors} we compare our proposed methods with its SO version, and observe that masking out all the \textit{background anchors} yields an overall improvement of $4.12\%$ and $5.92\%$ mAP, respectively. Hence, SADA reduces the impact of masking the \textit{background anchors} by $1.8\%$ mAP, indicating that our fine-grained alignment permits a better knowledge transfer to the target domain, mitigating the negative effect of these anchors. Find in the Supp. the complete ablation.
\begin{table}[t]
\centering
\footnotesize
    \renewcommand{\arraystretch}{0.9}
\setlength{\tabcolsep}{4pt} 
\resizebox{0.45\textwidth}{!}{%
\begin{tabular}{c|c|ccc|c|c}
\toprule
\textbf{Method} & \textbf{Mask \textit{bkg anchors}} & \multicolumn{3}{c|}{\textbf{mAP \{10,30,50\}\%}} & \textbf{Avg} & \textbf{Perf. gap} \\
\midrule
\multirow{2}{*}{Ours(src-only)} & & 30.03 & 26.62 & 17.79 & 24.81 & - \\
& \boldcheckmark & 35.25 & 33.71 & 23.25  & 30.73 & 5.92 \\
\midrule
\multirow{2}{*}{Ours(SADA)} & & 32.69 & 29.17 & 19.72 & 27.19 & - \\
 & \boldcheckmark & \textbf{35.70} & \textbf{34.09} & \textbf{24.15} & \textbf{31.31} & \textbf{4.12} \\
 \bottomrule
\end{tabular}}%
\vspace{-0.15cm}
\caption{Ablation of the effect of the \textit{background anchors} in the performance of the model on S3.}
\vspace{-0.45cm}
\label{tab:ablation_effect_background_anchors}
\end{table}

\noindent\textbf{Qualitative analysis.} We complement our quantitative results with a qualitative study. For this, we depict in Fig.~\ref{fig:video_qualitative_results} a segment visualization of S3 of our proposed method versus the chosen baseline models. In this visualization, we can observe that the Actionformer misses many of the segments in the shown video clip ignoring all the \textit{take} actions and mistaking a \textit{close} for a \textit{take}. Tridet performs better but misses all the \textit{take} actions in the first half of the video while worsening the boundary prediction of the last \textit{close}. \textit{SADA} improves Tridet by predicting the second \textit{take} and considerably improving the boundary of the last \textit{close} action.

\begin{figure}[t]
\centering
\includegraphics[width=\columnwidth]{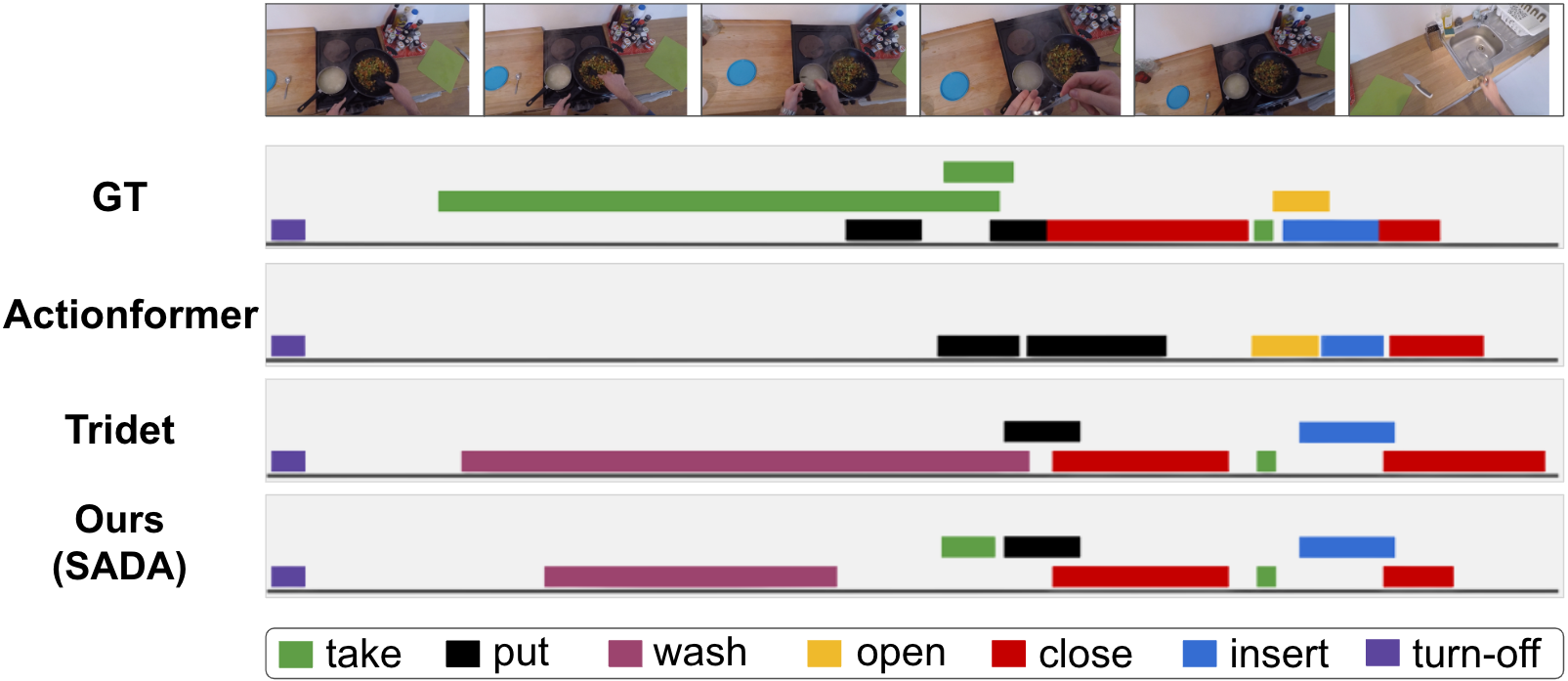}
\vspace{-0.6cm}
\caption{Visualization of the predicted segments of our method and the chosen set of source-only (SO). We include on top the ground-truth (GT) segments as a reference.}
\vspace{-0.3cm}
\label{fig:video_qualitative_results}
\end{figure}

\begin{figure}
\centering
\begin{subfigure}[b]{0.1\textwidth}
    \centering
    \includegraphics[width=2cm]{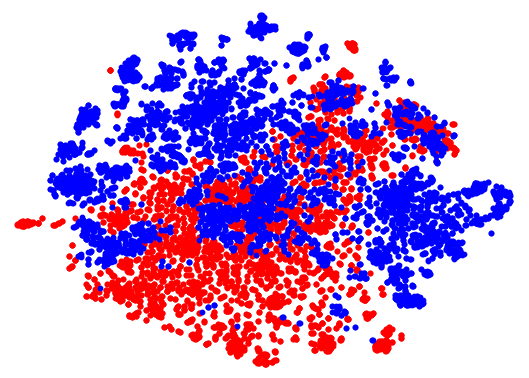}
    \caption{Class1(SO)\label{fig:image1}}
\end{subfigure}
\quad
\begin{subfigure}[b]{0.1\textwidth}
    \centering
    \includegraphics[width=2cm]{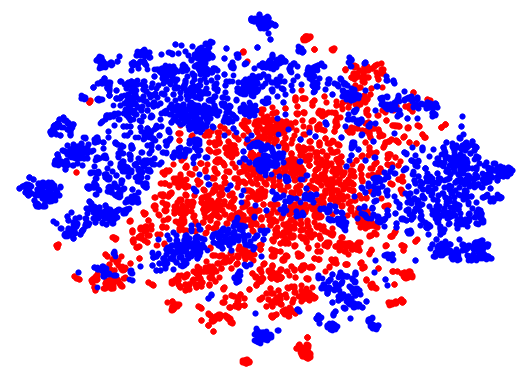}
    \caption{Class2(SO)\label{fig:image2}}
\end{subfigure}
\quad
\begin{subfigure}[b]{0.1\textwidth}
    \centering
    \includegraphics[width=2cm]{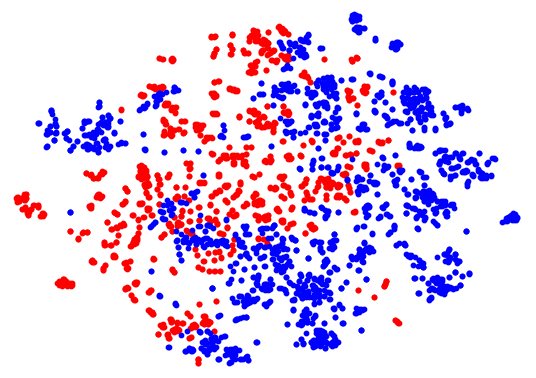}
    \caption{Class3(SO)\label{fig:image3}}
\end{subfigure}
\quad
\begin{subfigure}[b]{0.1\textwidth}
    \centering
    \includegraphics[width=2.1cm]{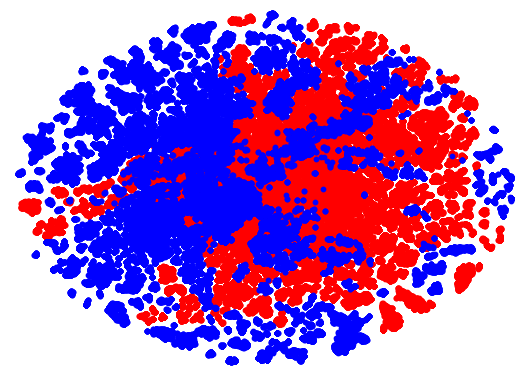}
    \caption{\hspace{0.25cm}Bkg(SO)\label{fig:image4}}
\end{subfigure}
\hfill\\
\begin{subfigure}[b]{0.1\textwidth}
    \centering
    \includegraphics[width=2cm]{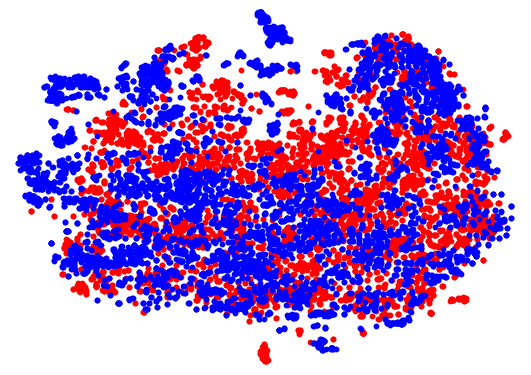}
    \caption{Class1(DA)\label{fig:image5}}
\end{subfigure}
\quad
\begin{subfigure}[b]{0.1\textwidth}
    \centering
    \includegraphics[width=2cm]{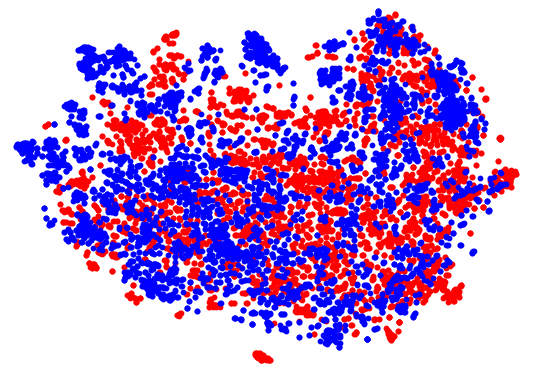}
    \caption{Class2(DA)\label{fig:image6}}
\end{subfigure}
\quad
\begin{subfigure}[b]{0.1\textwidth}
    \centering
    \includegraphics[width=2cm]{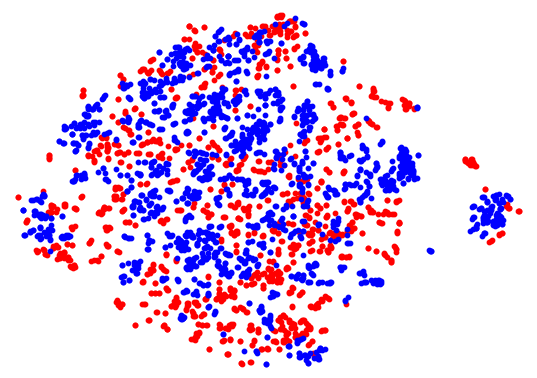}
    \caption{Class3(DA)\label{fig:image7}}
\end{subfigure}
\quad
\begin{subfigure}[b]{0.1\textwidth}
    \centering
    \includegraphics[width=2cm]{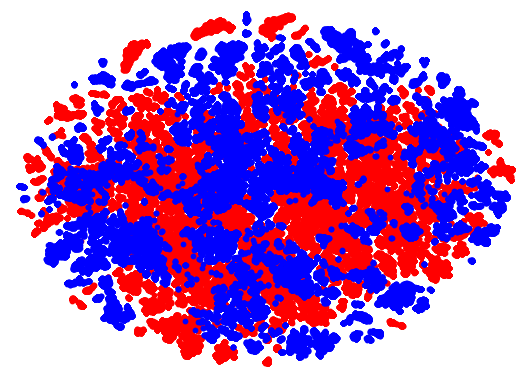}
    \caption{\hspace{0.25cm}Bkg(DA)\label{fig:image8}}
\end{subfigure}

\vspace{-0.15cm}
\caption{TSNE plots of the source-only (SO) variation of our model (top row) and our proposed domain adaptation model (DA) (bottom row). Find in the first 3 columns the TSNE plots of action classes 1 to 3 of the source (red) and target (blue) domain anchors. The last column shows the plot of the background anchors, so those not assigned to any GT label.}
\vspace{-0.4cm}
\label{fig:tsne_plots}
\end{figure}

Fig.~\ref{fig:tsne_plots} also shows the TSNE plots of the domain-invariant embeddings of \textit{SADA} with respect to learning source-only. Given that \textit{SADA} is an adversarial-based class-wise loss, we depict the plots of the 3 majority action classes (first 3 columns). Observe that SO (top row) yields clearly unaligned distributions with scarce to no overlap in the projected space. Our method, in contrast, presents a considerable distribution mix improving the alignment of class-wise distributions across domains. Given the anchor-based nature of our method, we have numerous \textit{background anchors} --i.e., not assigned to any GT label. As observed in the last column, these are also aligned by our method (see Eq. \ref{eq:bkg_loss}) therefore effectively aligning the entire data distributions but in a semantically sensitive way. See the Supp. for the complete study at different resolution levels.

\section{Conclusions}
In this work, we deal for the first time with Unsupervised Domain Adaptation on realistic Temporal Action Localization (TAL) scenarios. We propose a novel semantic adversarial loss that enables a more fine-grained distribution alignment compared to existing global-distribution-based approaches. Given the lack of suitable evaluation setups for this scenario, we propose a suite of 7 different benchmarks that provide a comprehensive assessment of the model performance across various domain shifts. These experiments indicate that our approach yields a considerable improvement over state-of-the-art methods, which we support with extensive quantitative and qualitative results.\\

\noindent \textbf{Ackwoledgements:} This work has been partially supported by the Spanish project PID2022-136436NB-I00 and by ICREA under the ICREA Academia programme.

%
%
\bibliographystyle{splncs04}
\bibliography{main}

\newpage
\newpage
\renewcommand*{\thesection}{\Alph{section}}
\renewcommand*{\thefigure}{\Alph{figure}}
\renewcommand*{\thetable}{\Alph{table}}
\setcounter{section}{0}
\setcounter{table}{0}
\setcounter{figure}{0}

\clearpage
\noindent {\Large \textbf{Supplementary Material}\par}
\vspace{0.5cm}

In this supplementary material, we first provide the implementation details needed to reproduce our work (see Sec.~\ref{sec:appendix_implementation_details}). We then extend the description of Sec. 4.1 regarding our 7 proposed benchmarking scenarios. In Sec.~\ref{sec:appendix_unsupervised_domain_adaptation_baselines} we describe in detail our chosen UDA baselines. We then complement the provided ablation studies from Sec. 4.3 (see Sec.~\ref{sec:appendix_ablation_da} and Sec.~\ref{sec:appendix_ablation_variants_sada}). In Sec.~\ref{sec:appendix_effect_of_lambda_hyperparam} we ablate over the importance of properly tunning the $\lambda_l$ hyperparameter and in Sec.~\ref{sec:appendix_study_of_the_class_embedding} we provide empirical evidences that justify the choice of using learnable class embeddings. In Sec.~\ref{sec:appendix_per_class} we provide a comprehensive study of the per-class performance of our model on multiple scenarios. In Sec.~\ref{sec:appendix_ablation_negative_anchors} we ablate over the capacity of SADA to mitigate the potential negative impact of \textit{background anchors}, and provide empirical evidences of the need of further research to continue alleviating the degradation in performance that they induce. Finally, in Sec.~\ref{sec:appendix_qualitative_results} we provide complementary qualitative results, including further segment visualizations (see Sec.~\ref{sec:appendix_segment_visualization}) and TSNE plots (see Sec.~\ref{sec:appendix_extended_tsne_analysis}). 
\section{Implementation details}\label{sec:appendix_implementation_details}
In this section we include all the relevant implementation details that ensure the proper reproducibility of \textit{SADA} in the different setups that we present in this paper. Note that we report only the hyperparameters of the best model only, which were obtained through a Bayesian optimization process to minimize the downstream loss --i.e., not including the domain adaptation losses-- of the validation split of the Source domain. We follow this approach since by definition we do not have any label information on the Target domain. Nevertheless, we hypothesize that this is a suitable strategy as the goal of our adaptation loss is to \textit{bring both domains closer}. Thus, a set of hyperparameters that performs well on the Source domain should similarly do so on the Target domain. We mitigate the presence of overfitting by using early stopping and a weight decay with a value of $0.05$.

All the models that we present are implemented using PyTorch-2.0, CUDA  12.4 and trained for 50 epochs using AdamW~\cite{loshchilov2017decoupled} optimizer with a learning rate of $1e^{-4}$ with cosine decay and a warm-up phase of $5$ epochs on one single NVIDIA GeForce RTX 3090. The training process in all our experiments took between 12 and 24 hours.

Architecturally, all our models define the main feature extractor as a 6-level SGP feature pyramid~\cite{shi2023tridet} that follows a max pooling strategy and uses a downsampling rate of $2$ and an internal feature dimensionality of $1024$. We model the classification and localization heads as a sequence of three 1D-CNNs with a kernel size of $3$. These heads are shared across embeddings of different resolution levels. Finally, we introduce level-wise domain discriminators designed as a multi-layer perceptron (MLP) of depth $2$ and width $512$. Find in Tab. \ref{tab:implementation_details_sada} the other relevant scenario-dependent hyperparameters related to our main contribution, the \textit{SADA} loss. The rest of the hyperparameters follow the original implementation of Tridet~\cite{shi2023tridet}.

\begin{table*}[t]
    \centering
    \resizebox{0.6\textwidth}{!}{%
    \begin{tabular}{c|c|c|ccccccc|ccc}
        \toprule
        Dataset & Scenario & Backbone & $\lambda_1$ & $\lambda_2$ & $\lambda_3$ & $\lambda_4$ & $\lambda_5$ & $\lambda_6$ & $\alpha$ & $\lambda_{reg}$ & $\lambda_{cls}$ & $\lambda_{sada}$\\ 
        \midrule
        \multirow{6}{*}{EK100} & S1 & SF & 0.7 & 0.3 & 0.9 & 0.8 & 0.7 & 0.0 & 0.3 & 1.0 & 1.0 & 3.0 \\ 
        & S2 & SF & 0.7 & 0.6 & 1.0 & 0.7 & 0.2 & 0.2 & 0.3 & 1.0 & 1.0 & 3.0 \\ 
        & S3 & SF & 0.4 & 0.8 & 0.7 & 0.7 & 0.9 & 0.6 & 0.6 & 1.0 & 1.0 & 1.0 \\
        & S4 & SF & 0.4 & 0.9 & 0.1 & 0.4 & 0.1 & 0.6 & 0.5 & 1.0 & 1.0 & 0.5 \\
        & S5 & SF & 1.0 & 0.4 & 0.8 & 0.7 & 0.9 & 0.6 & 0.5 & 1.0 & 1.0 & 2.0 \\
        & S6 & SF & 0.6 & 0.1 & 1.0 & 0.6 & 0.2 & 0.4 & 0.3 & 1.0 & 1.0 & 2.0 \\ 
        \midrule
        CharadesEgo & - & I3D & 0.4 & 0.8 & 0.7 & 0.7 & 0.9 & 0.6 & 0.6 & 1.0 & 1.0 & 1.0 \\ 
        \bottomrule
    \end{tabular}}
    \vspace{0.3cm}
    \caption{Summary of the main hyperparameters detailed for each of the presented benchmarks. Here I3D refers to \cite{carreira2017quo} and SF to \cite{feichtenhofer2019slowfast}.}
    \label{tab:implementation_details_sada}
\end{table*}

During training, \textit{SADA} first extracts video features using a frozen video backbone. For all the scenarios based on EpicKitchens100 this relies on a Slowfast~\cite{carreira2017quo} video backbone pre-trained on Kinetics~\cite{kay2017kinetics}. This backbone is fed with raw videos from EpicKitchens100~\cite{Damen2022RESCALING} with a rate of $30$ FPS, a feature stride of $16$, and a maximum length of $2304$ ensured by either padding -- for shorter videos -- or random cropping. Alternatively, in the case of CharadesEgo\cite{sigurdsson2018charades} we rely on a frozen I3D video backbone pre-trained on ImageNet\cite{deng2009imagenet}. Our model then shuffles the pre-extracted features of the source and target domain videos, respectively. It then feeds a batch of $2$ videos of each of the domains, resulting in a batch of $4$ videos per training iteration. Note that depending on the scenario under study, one of the domains may have more data than the other. For this reason, in each epoch we repeat the smaller domain until the other finishes, ensuring that all the data of each of the domains is leveraged at least once.

During inference, we follow a similar approach to \cite{zhang2022actionformer, shi2023tridet} and define an exponential moving average of the trained model, which we update every iteration with an exponential decay of 0.999. Moreover, given the excess of final predictions, we use the standard SoftNMS~\cite{bodla2017soft} with an IOU threshold of 0.1, a minimum score threshold of 0.001, and a sigma value of 0.4.

\section{Extended benchmark description}\label{sec:appendix_extended_banchmarks}
In this section we describe in more depth the 7 different benchmarking scenarios that we propose in Sec. \ref{sec:experimentation_experimental_setup}. Concretely, in Sec. \ref{sec:appendix_epic} we expand on the 6 different setups that build on the EpicKitchens100 dataset. For this, in Sec.~\ref{sec:appendix_description_epic} we first briefly describe EpicKitchens100~\cite{Damen2022RESCALING}, the base of all our benchmarks. We then illustrate in Sec.~\ref{appendix:acquisition_shift_description} and Sec.~\ref{sec:appendix_appearance_shift_description} the 2 different domain shifts that we identify, which we complement with a more detailed description of the specific scenarios derived from each of them. In Sec.~\ref{sec:appendix_study_benchmark_statistics} we also include a detailed set of statistics for each of the proposed setups. We proceed similarly for the scenario based on Charades-Ego (see Sec.~\ref{sec:appendix_charadesego}). This includes a brief description of the dataset and a detailed explanation of the domain adaptation benchmark that we propose in this work based on this dataset.

\subsection{EpicKitchens100 scenarios}\label{sec:appendix_epic}
\textbf{Description:}\label{sec:appendix_description_epic}
EpicKitchens100~\cite{Damen2022RESCALING} is a large-scale ego-centric dataset that has recently gained a lot of attraction in the Temporal Action Localization community to test for challenging detection scenarios. This dataset extends the previous version of the dataset, namely EpicKitchens55~\cite{Damen2018EPICKITCHENS}. This dataset contains videos of 32 subjects performing daily cooking activities -- e.g., wash, put, open or close -- in different kitchen environments. This results in 100 hours of non-scripted videos containing 20M frames annotated with 90K action segments.

One of the key challenges of this dataset is the presence of 97 different actions, forming a very long tail action distribution. This inherent characteristic is however an issue that falls beyond the scope of this paper, given the important additional challenges that this poses in existing domain adaptation-based methods. For this reason, in all our experiments we consider only the 10 majority classes -- i.e., \textit{take}, \textit{put}, \textit{wash}, \textit{open}, \textit{close}, \textit{insert}, \textit{turn-on}, \textit{cut}, \textit{turn-off} and \textit{pour}. As shown in Fig. \ref{fig:histogram_class_distribution} this pruning keeps most of the labeled action segments of the original dataset, concretely keeping up to $80\%$ of it (marked in blue in the histogram). Moreover, we find that this design decision does not diminish in a relevant way the challenging nature of this task. Proof of this is that the currently best-performing methods on this dataset -- i.e., Actionformer~\cite{zhang2022actionformer} and Tridet~\cite{shi2023tridet}-- still attained mAP metrics inferior to $30\%$ in all our proposed scenarios. 

\begin{figure}
\centering
\includegraphics[width=0.48\textwidth]{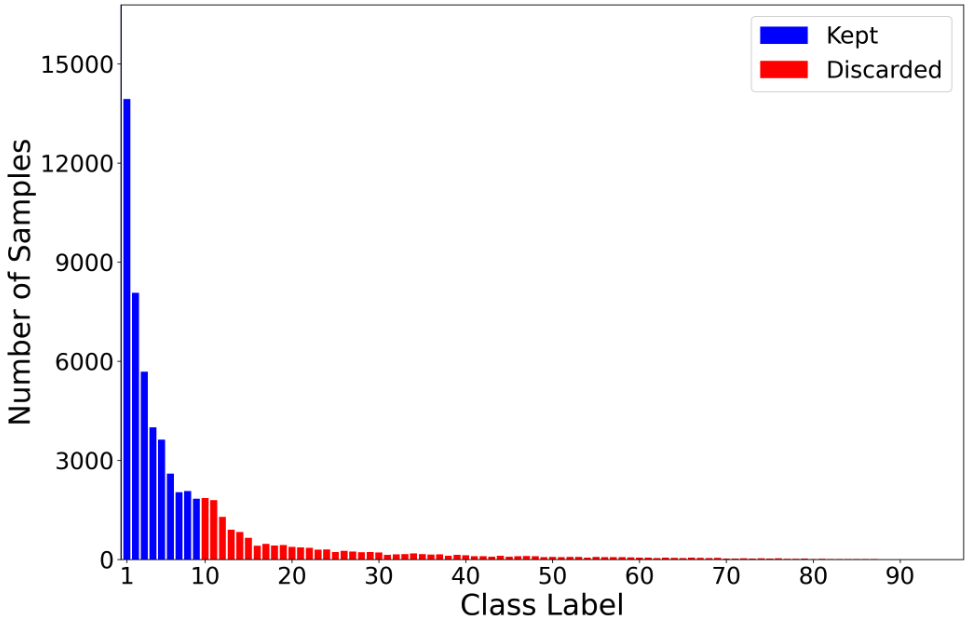}
\caption{Histogram of the number of GT action segments of every class. We also depict in blue the 10 majority classes --which we keep in our setups-- while leaving as red the remaining class that we discard to avoid the long-tail action distribution problem.}
\label{fig:histogram_class_distribution}
\vspace{-0.3cm}
\end{figure}

\noindent\textbf{Studying acquisition shifts:}\label{appendix:acquisition_shift_description}
The first domain shift that we identify in the EpicKitchens100~\cite{Damen2022RESCALING} is what we call the \textit{acquisition shift}. This shift refers to the changes induced by differences in the acquisition conditions of the videos. In the case of EpicKitchens100~\cite{Damen2022RESCALING} this results from the extension of the original dataset EpicKitchens55~\cite{Damen2018EPICKITCHENS}. The original dataset, concretely, was formed by 55 hours of non-scripted videos, and nearly 40K action segments. Hence, we find that this presents a suitable setup to define one domain as the \textit{old} videos -- recorded in the original EpicKitchens55~\cite{Damen2018EPICKITCHENS} -- and the other formed by the \textit{new} videos -- recorded for the extended version. As argued by \cite{Damen2022RESCALING} this results in several important domain gaps that are captured by these data splits:
\begin{itemize}[nosep]
    \item \textbf{Changes in the acquisition devices:} The \textit{new} videos that were recorded during the extension relied on a newer camera device. Importantly, this camera incorporates camera stabilization techniques. Fig.~\ref{fig:appendix_acquisition_shift} visually depicts the improvement in the stability of \textit{new} over \textit{old} videos.
    \item \textbf{Lighting conditions:} Given the changes in the hour of the recording, these domains also present differences in the lighting conditions of the videos.
\end{itemize}

\begin{figure*}[t]
    \centering
    \begin{subfigure}[b]{\textwidth}
        \centering
        \includegraphics[width=\textwidth]{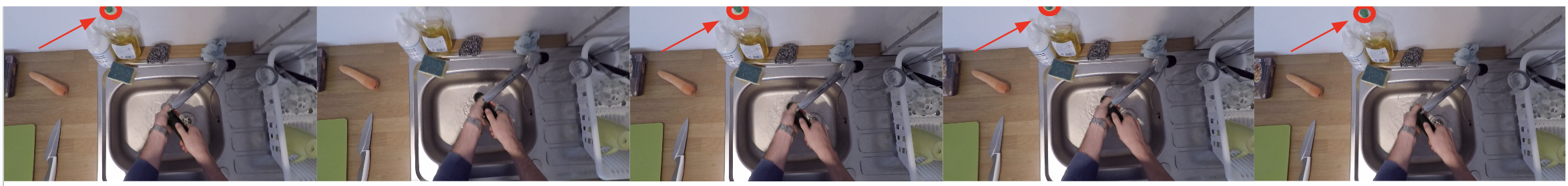}
        \caption{\textit{Old} videos}
        \label{fig:old_kitchens}
    \end{subfigure}
    \\
    \begin{subfigure}[b]{\textwidth}
        \centering
        \includegraphics[width=\textwidth]{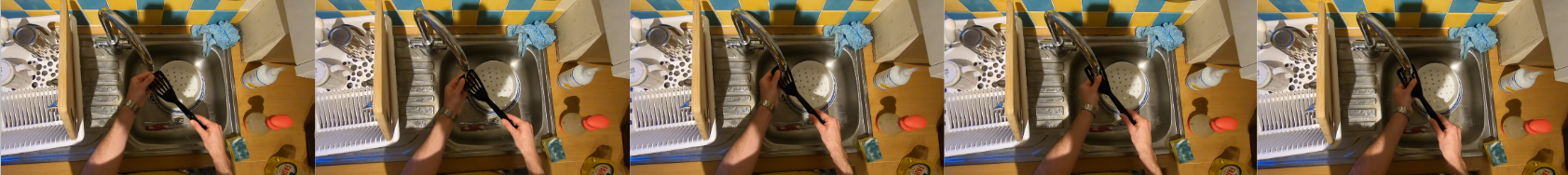}
        \caption{\textit{New} videos}
        \label{fig:new_kitchens}
    \end{subfigure}
    \caption{Visualization of the differences in camera stability between \textit{old} and \textit{new} videos. These images were obtained with differences of 3 frames, which correspond to a period of 0.1 seconds. We highlight with a red circle in \ref{fig:old_kitchens} a reference in the images to visualize the high instability of the video. }
    \label{fig:appendix_acquisition_shift}
\end{figure*}

\noindent\textbf{Studying appearance shifts:}\label{sec:appendix_appearance_shift_description}
Another important domain shift that we are interested in capturing is one induced by changes in the background. In this regard, there are several possibilities for this, but we find that the vast majority allow only a vague intuition of the domain shift that is under study. For instance, \cite{Damen2022RESCALING} argues that in the extension, several kitchens might have some changes in the furniture or the order of the main kitchen utensils. But, \textit{which kitchens really contain these changes?} Not understanding this issue in sufficient depth results in an obscure evaluation, which we try to avoid in this work.

For this reason, we identify a clear, understandable domain gap. This is the color of the kitchen counters. This is an essential part of the background information, and thus, we hypothesize that adapting to this factor is also critical to obtain high-performing models. In this regard, we split the data into three different domains: \textit{dark}, \textit{white}, and \textit{other} types of kitchen counters. This provides a clear domain gap that we can visualize and thus, understand (see Fig. \ref{fig:appearance_shift_kitchens}). 

Intuitively, this permits to define the following 2 scenarios. Firstly, we can define a scenario for the black-counter kitchens as the source domain, while the rest are the target domain. Similarly, we define another scenario where the white-counter kitchens are the source domain, while the rest are the target domain. Notice, however, that this mixes acquisition conditions as we do not differentiate \textit{old} and \textit{new} videos. To ensure that this does not impede a clear understanding of the scenario, we consider \textit{old} and \textit{new} videos, independently. This results in our 4 final scenarios to measure appearance shifts.

All in all, we stress that all our proposed scenarios provide sufficiently even splits to design single-domain setups where we can ensure that there is no easy generalization of the target domain by means of the source training data (as argued in Sec. \ref{sec:experimentation_experimental_setup}).

\begin{figure*}[t]
    \centering
    \begin{subfigure}[b]{0.3\textwidth}
        \centering
        \includegraphics[width=\textwidth]{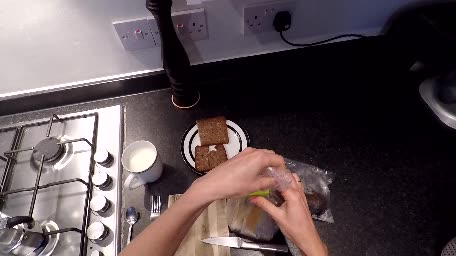}
        \caption{Dark kitchen}
        \label{fig:dark_kitchen}
    \end{subfigure}
    \hfill
    \begin{subfigure}[b]{0.3\textwidth}
        \centering
        \includegraphics[width=\textwidth]{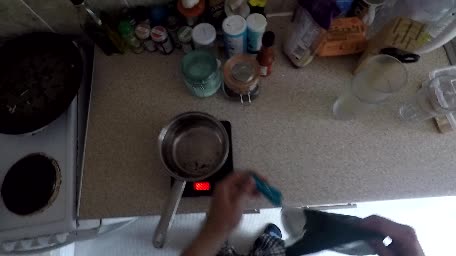}
        \caption{White kitchen}
        \label{fig:white_kitchen}
    \end{subfigure}
    \hfill
    \begin{subfigure}[b]{0.3\textwidth}
        \centering
        \includegraphics[width=\textwidth]{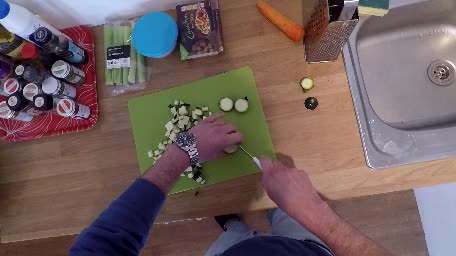}
        \caption{Other kitchen}
        \label{fig:brown_kitchen}
    \end{subfigure}
    \caption{Visualization of the 3 different appearance-based domain shifts resulting from the split of dark-counter kitchens, white-counter kitchens, and finally all the other kitchens.}
    \label{fig:appearance_shift_kitchens}
\end{figure*}

\noindent\textbf{Study of the benchmark statistics:}\label{sec:appendix_study_benchmark_statistics}
In Sec. \ref{sec:experimentation_experimental_setup} we propose a set of 6 new scenarios to perform UDA on egocentric videos based on EpicKitchens100~\cite{Damen2022RESCALING}. To do so we rely on two main factors to split the original data from EpicKitchens100~\cite{Damen2022RESCALING} into domains: color of the kitchen counter and year of acquisition --i.e., if the data was already part of EpicKitchens55~\cite{Damen2018EPICKITCHENS}. In this regard, in Tab. \ref{tab:domain_splits_statistics} we detail the relevant statistics according to these splits where we present the number of videos, the number of segments, the average length of these segments, and finally the class-wise number of segments to depict the overall distribution of actions.

Given this data splits, we define 6 different scenarios according to the description of Sec. \ref{sec:experimental_results}. In this respect, in Tab. \ref{tab:scenario_source_statistics} and Tab. \ref{tab:scenario_target_statistics} we present the aforementioned statistics for the source and target domain of each scenario, respectively. Observe that thanks to our careful experimental setup, we are able to define scenarios with considerably large source and target domains. This contrasts with other experimental setups like that proposed by \cite{wei2023unsupervised} which define the source and target domain of a scenario as the videos of a single kitchen, respectively. We find that this, in consequence, yields domains with only 15-28 videos, which we deem insufficient to train state-of-the-art large architectures.

\begin{table*}[t]
\centering
\resizebox{0.85\textwidth}{!}{%
\begin{tabular}{c|c|c|c|c|c|cccccccccc}
\toprule
\textbf{Color} & \textbf{Year} & \textbf{Split} & \textbf{\# vids} & \textbf{\# segs} &  \textbf{Avg len (s)} & \textbf{c1} & \textbf{c2} & \textbf{c3} & \textbf{c4} & \textbf{c5} & \textbf{c6} & \textbf{c7} & \textbf{c8} & \textbf{c9} & \textbf{c10} \\
\midrule
\multirow{4}{*}{Dark} & \multirow{2}{*}{55} & Train & 126 & 8336 & 3.11 & 1982 & 1611 & 1208 & 868 & 561 & 768 & 315 & 492 & 243 & 288 \\
& & Val & 35 & 2944 & 3.05 & 699 & 603 & 446 & 273 & 194 & 271 & 131 & 124 & 92 & 111 \\
\cline{2-16}
& \multirow{2}{*}{100} & Train & 45 & 5487 & 2.34 & 1576 & 1209 & 672 & 429 & 314 & 453 & 262 & 186 & 226 & 160 \\
& & Val & 13 & 1504 & 1.95 & 430 & 320 & 177 & 127 & 80 & 113 & 98 & 11 & 69 & 79 \\
\midrule
\multirow{4}{*}{White} & \multirow{2}{*}{55} & Train & 124 & 9995 & 2.95 & 2453 & 2398 & 1739 & 963 & 658 & 439 & 477 & 321 & 352 & 195 \\
& & Val & 31 & 1595 & 3.75 & 385 & 313 & 184 & 189 & 111 & 101 & 78 & 120 & 56 & 58 \\
\cline{2-16}
& \multirow{2}{*}{100} & Train & 49 & 6895 & 2.26 & 2236 & 1889 & 818 & 579 & 417 & 87 & 313 & 109 & 245 & 202 \\
& & Val & 19 & 2328 & 2.79 & 705 & 553 & 377 & 184 & 135 & 67 & 107 & 73 & 83 & 44 \\
\midrule
\multirow{4}{*}{Brown} & \multirow{2}{*}{55} & Train & 30 & 3013 & 3.13 & 852 & 723 & 492 & 277 & 229 & 141 & 87 & 98 & 63 & 141 \\
& & Val & 7 & 640 & 2.95 & 177 & 135 & 55 & 78 & 67 & 43 & 18 & 21 & 16 & 30 \\
\cline{2-16}
& \multirow{2}{*}{100} & Train & 10 & 1383 & 2.12 & 422 & 330 & 228 & 108 & 92 & 47 & 55 & 18 & 52 & 31 \\
& & Val & 2 & 663 & 2.13 & 235 & 172 & 54 & 43 & 37 & 35 & 19 & 41 & 16 & 11 \\
\midrule
\multirow{4}{*}{Other} & \multirow{2}{*}{55} & Train & 62 & 4200 & 3.53 & 1162 & 797 & 389 & 609 & 314 & 360 & 134 & 198 & 102 & 135 \\
& & Val & 17 & 331 & 3.81 & 91 & 59 & 30 & 47 & 28 & 36 & 5 & 19 & 6 & 10 \\
\cline{2-16}
& \multirow{2}{*}{100} & Train & 46 & 6747 & 1.72 & 2113 & 1593 & 497 & 599 & 496 & 586 & 300 & 146 & 265 & 152 \\
& & Val & 10 & 2031 & 1.97 & 594 & 528 & 204 & 186 & 143 & 76 & 82 & 51 & 73 & 94 \\
\bottomrule
\end{tabular}}
\vspace{0.3cm}
\caption{Analysis of the different proposed domains, with their corresponding data splits.}
\label{tab:domain_splits_statistics}
\end{table*}

\begin{table*}[t]
\centering
\resizebox{\textwidth}{!}{%
\begin{tabular}{c|c|c|c|c|c|cccccccccc}
\toprule
\textbf{Scenario} & \textbf{Domain} & \textbf{Split} & \textbf{\# vids} & \textbf{\# segs} & \textbf{Avg len (s)} & \textbf{c1} & \textbf{c2} & \textbf{c3} & \textbf{c4} & \textbf{c5} & \textbf{c6} & \textbf{c7} & \textbf{c8} & \textbf{c9} & \textbf{c10} \\
\midrule
\multirow{2}{*}{S1} & \multirow{2}{*}{Dark(55)} & Train & 126 & 8336 & 3.11 & 1982 & 1611 & 1208 & 868 & 561 & 768 & 315 & 492 & 243 & 288 \\
& & Val & 35 & 2944 & 3.05 & 699 & 603 & 446 & 273 & 194 & 271 & 131 & 124 & 92 & 111 \\
\midrule
\multirow{2}{*}{S2} & \multirow{2}{*}{White(55)} & Train & 124 & 9995 & 2.95 & 2453 & 2398 & 1739 & 963 & 658 & 439 & 477 & 321 & 352 & 195 \\
& & Val & 31 & 1595 & 3.75 & 385 & 313 & 184 & 189 & 111 & 101 & 78 & 120 & 56 & 58 \\
\midrule
\multirow{2}{*}{S3} & \multirow{2}{*}{Dark(100)} & Train & 45 & 5487 & 2.34 & 1576 & 1209 & 672 & 429 & 314 & 453 & 262 & 186 & 226 & 160 \\
& & Val & 13 & 1504 & 1.95 & 430 & 320 & 177 & 127 & 80 & 113 & 98 & 11 & 69 & 79 \\
\midrule
\multirow{2}{*}{S4} & \multirow{2}{*}{White(100)} & Train & 49 & 6895 & 2.26 & 2236 & 1889 & 818 & 579 & 417 & 87 & 313 & 109 & 245 & 202 \\
& & Val & 19 & 2328 & 2.79 & 705 & 553 & 377 & 184 & 135 & 67 & 107 & 73 & 83 & 44 \\
\midrule
\multirow{2}{*}{S5} & \multirow{2}{*}{All(55)} & Train & 342 & 25634 & 3.12 & 5529 & 6449 & 1762 & 2717 & 1708 & 1013 & 3828 & 760 & 1109 & 759 \\
& & Val & 90 & 5510 & 3.29 & 1352 & 1110 & 451 & 587 & 400 & 170 & 715 & 209 & 284 & 232 \\
\midrule
\multirow{2}{*}{S6} & \multirow{2}{*}{All(100)} & Train & 150 & 20512 &  2.10 & 930 & 6347 & 1319 & 788 & 1173 & 5021 & 1715 & 545 & 459 & 2215 \\
& & Val & 44 & 6526 & 2.27 & 1573 & 540 & 395 & 306 & 1964 & 241 & 228 & 291 & 812 & 176 \\
\bottomrule
\end{tabular}}
\vspace{0.3cm}
\caption{Analysis of the Source domains of each of the proposed scenarios.}
\label{tab:scenario_source_statistics}
\end{table*}

\begin{table*}[t]
\centering
\resizebox{\textwidth}{!}{%
\begin{tabular}{c|c|c|c|c|c|cccccccccc}
\toprule
\textbf{Scenario} & \textbf{Domain} & \textbf{Split} & \textbf{\# vids} & \textbf{\# segs} & \textbf{Avg len (s)} & \textbf{c1} & \textbf{c2} & \textbf{c3} & \textbf{c4} & \textbf{c5} & \textbf{c6} & \textbf{c7} & \textbf{c8} & \textbf{c9} & \textbf{c10} \\
\midrule
\multirow{2}{*}{S1} & \multirow{2}{*}{All-Dark(55)} & Train & 216 & 17298 & 3.12 & 3918 & 4467 & 1201 & 1849 & 940 & 698 & 2620 & 517 & 617 & 471 \\
& & Val & 55 & 2566 & 3.56 & 653 & 507 & 180 & 314 & 206 & 78 & 269 & 98 & 160 & 101 \\
\midrule
\multirow{2}{*}{S2} & \multirow{2}{*}{All-White(55)} & Train & 218 & 15639 & 3.23 & 3131 & 3996 & 1104 & 1754 & 1269 & 536 & 2089 & 408 & 788 & 564 \\
& & Val & 59 & 3915 & 3.10 & 967 & 797 & 350 & 398 & 289 & 114 & 531 & 151 & 164 & 154 \\
\midrule
\multirow{2}{*}{S3} & \multirow{2}{*}{All-Dark(100)} & Train & 105 & 15025 & 2.02 & 668 & 4771 & 1005 & 562 & 720 & 3812 & 1286 & 385 & 273 & 1543 \\
& & Val & 31 & 5022 & 2.38 & 1253 & 413 & 315 & 208 & 1534 & 172 & 149 & 178 & 635 & 165 \\
\midrule
\multirow{2}{*}{S4} & \multirow{2}{*}{All-White(100)} & Train & 101 & 13617 & 2.01 & 617 & 4111 & 902 & 543 & 1086 & 3132 & 1136 & 343 & 350 & 1397 \\
& & Val & 25 & 4198 & 1.99 & 1020 & 356 & 260 & 199 & 1259 & 158 & 184 & 224 & 435 & 103\\
\midrule
\multirow{2}{*}{S5} & \multirow{2}{*}{All(100)} & Train & 150 & 20512 & 2.10 & 930 & 6347 & 1319 & 788 & 1173 & 5021 & 1715 & 545 & 459 & 2215 \\
& & Val & 44 & 6526 & 2.27 & 1573 & 540 & 395 & 306 & 1964 & 241 & 228 & 291 & 812 & 176 \\
\midrule
\multirow{2}{*}{S6} & \multirow{2}{*}{All(55)} & Train & 342 & 25634 & 3.12 & 5529 & 6449 & 1762 & 2717 & 1708 & 1013 & 3828 & 760 & 1109 & 759 \\
& & Val & 90 & 5510 & 3.29 & 1352 & 1110 & 451 & 587 & 400 & 170 & 715 & 209 & 284 & 232 \\
\bottomrule
\end{tabular}}
\vspace{0.3cm}
\caption{Analysis of the Target domains of each of the proposed scenarios.}
\label{tab:scenario_target_statistics}
\end{table*}

\subsection{CharadesEgo scenario}\label{sec:appendix_charadesego}
\noindent\textbf{Description:}\label{sec:appendix_description_charadesego}
Charades-Ego\cite{sigurdsson2018charades} is a dataset that was originally proposed in order to enable the transfer of knowledge from third-view videos to their corresponding first-person videos --a.k.a., egocentric videos. Concretely, they present a set of 4000 paired videos --i.e., 8000 videos in total-- involving 112 people with an average length of 31.2 seconds. Importantly, this dataset follows the methodology proposed by \cite{sigurdsson2016hollywood} and asks each of the users to record two videos: 1) Recording the action from a third-person perspective based on a pre-defined script 2) Perform the same action while recording it with a camera fixed in the forehead. Overall, this dataset encompasses a set of 157 complex actions --e.g., \textit{holding some clothes}, \textit{Taking a book from somewhere}.

\noindent\textbf{Third-to-egocentric setup:}\label{sec:appendix_third_to_egocentric}
Based on this dataset, we are the first to present a benchmark on domain adaptation for Temporal Action Localization. Our presented setup, concretely, explores the adaptation of third-person videos to egocentric ones. This sometimes involves an extreme domain shift as depicted in Fig.~\ref{fig:appendix_charadesego_visualization}. In fact we argue this domain shift can be so extreme that would not be suitable for image-based model benchmarking given that one of the views might lack necessary information that is only contained in the other view. For instance, \textit{a person awekens in bed} could be indiscernible from \textit{a person looking at the ceiling} if only one single frame is considered. Nevertheless, our proposed benchmark deals with video-inputs, which should provide sufficient context from the other frames to infer the true action, making this challenging setup more feasible.

To be more precise on our proposed benchmarking setup, we follow the work of \cite{zhang2022audio} and define the third-person videos as Source domain and the egocentric ones as Target domain. This setup presents an important limitation that we find is critical to make this benchmark viable in this complex task. Concretely, observe that class actions are mostly defined as NLP descriptions, which creates very concrete actions, making it cumbersome to adapt classes across domains. We argue that one of the main reasons is the lack of sufficient data for each of the actions. Consequently, we rely on the originally extended annotations from \cite{sigurdsson2018charades} and propose to use instead the \textit{verbs} of each of the corresponding action labels. This is, for a given action \textit{holding some clothes}, we consider this action with its verb \textit{hold}. One final consideration is that this still results in underrepresented \textit{verb} classes which we fix following a similar approach to that presented by \cite{wei2023unsupervised} and consider only the 10 majority classes being \textit{take}, \textit{put}, \textit{sit}, \textit{walk}, \textit{hold}, \textit{stand}, \textit{open}, \textit{drink}, \textit{smile} and \textit{laugh}. This still results in a very challenging setup as observed in Tab. \ref{tab:charadesego_results}, which indicates that none of the state-of-the-art methods tested attain more than $25\%$ mAP. Moreover, this setup retains $70\%$ of the original annotations.

\begin{figure*}
    \centering
    \begin{subfigure}[b]{0.475\textwidth}
        \centering
        \includegraphics[width=0.85\textwidth]{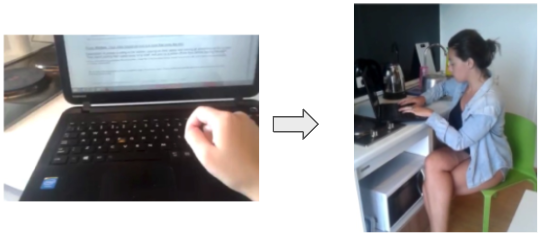}
        \caption{"A person is typing on a laptop"}
        \label{fig:charadesego_scenario1}
    \end{subfigure}
    \vline
    \hspace{0.1cm}
    \begin{subfigure}[b]{0.475\textwidth}
        \centering
        \includegraphics[width=\textwidth]{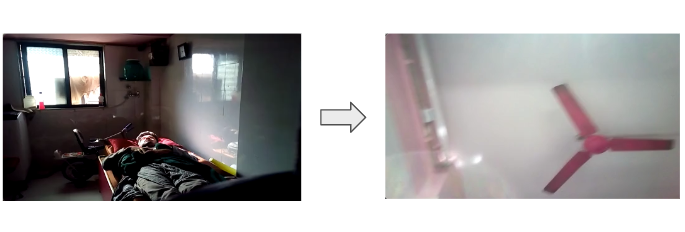}
        \caption{"A person awakens in bed."}
        \label{fig:charadesesgo_scenario2}
    \end{subfigure}
    \newline
    \begin{subfigure}[b]{0.475\textwidth}
        \centering
        \includegraphics[width=0.85\textwidth]{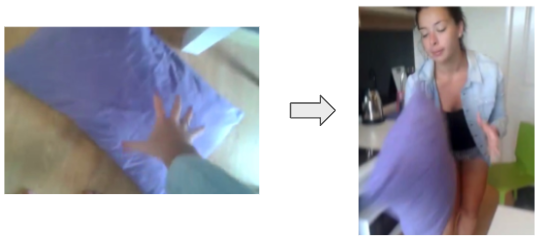}
        \caption{"A person picks up a pillow"}
        \label{fig:charadesego_scenario3}
    \end{subfigure}
    \vline
    \hspace{0.1cm}
    \begin{subfigure}[b]{0.475\textwidth}
        \centering
        \includegraphics[width=1\textwidth]{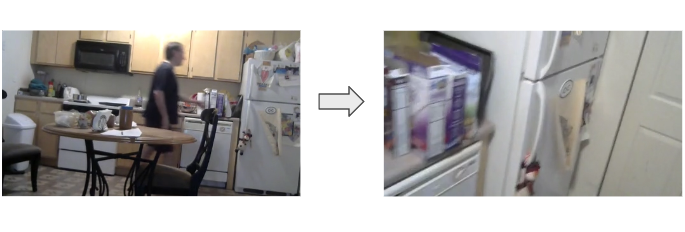}
        \caption{"A person pours a drink from the fridge"}
        \label{fig:charadesesgo_scenario4}
    \end{subfigure}
    \newline
        \begin{subfigure}[b]{0.475\textwidth}
        \centering
        \includegraphics[width=0.85\textwidth]{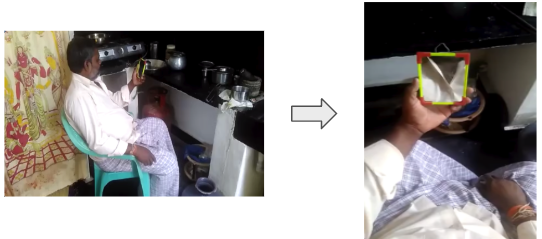}
        \caption{"A person is sitting in the pantry holding a mirror. "}
        \label{fig:charadesego_scenario5}
    \end{subfigure}
    \vline
    \hspace{0.2cm}
    \begin{subfigure}[b]{0.48\textwidth}
        \centering
        \includegraphics[width=\textwidth]{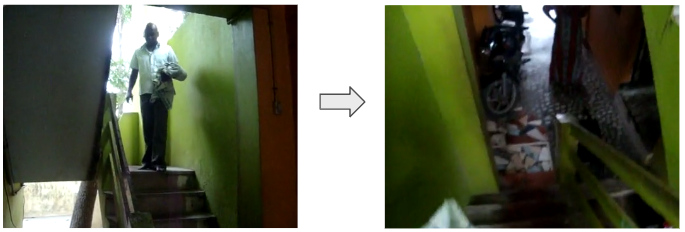}
        \caption{"A person is running down the stairs holding a blanket"}
        \label{fig:charadesesgo_scenario6}
    \end{subfigure}

    \caption{Comparison in two different scenarios between the egocentric perspective (left) and the third-person one (right).}
    \label{fig:appendix_charadesego_visualization}
    
\end{figure*}
\section{Unsupervised domain adaptation baselines}\label{sec:appendix_unsupervised_domain_adaptation_baselines}
In this section, we describe in greater detail the different baselines that we use in Sec. \ref{sec:ablation_studies} to establish a fair comparison with our work, \textit{SADA}. Concretely, we first describe the adaptations to our architecture of multiple UDA methods like DANN~\cite{ganin2015unsupervised}, ADDA~\cite{tzeng2017adversarial}, WDGRL~\cite{shen2018wasserstein}, MSTN~\cite{xie2018learning}, FGDA~\cite{gao2021gradient} or DRDA~\cite{huang2023discriminative}. We then describe the adaptation of the closest related work to \textit{SADA}, SSTDA~\cite{chen2020action} and for comparison purposes, the adaptation of one of the latest video action recognition methods, TranSVAE~\cite{wei2024unsupervised} to the task of Temporal Action Localization.

\textbf{DANN~\cite{ganin2015unsupervised}:} This work pioneered the use of adversarial losses to devise methods that learn domain invariant representations. They do so by proposing a domain classifier $D:\mathbb{R}^F \rightarrow \{0,1\}$ that learns to identify the domain that each of the feature embeddings belongs to. To integrate this proposal into a multi-resolution architecture we define a level-wise domain classifier $D_l:\mathbb{R}^F\rightarrow \{0,1\}$. Then, at a given level $l$, we apply the following adaptation loss on the feature representations of source domain $Z_l^\mathcal{S}\in\mathbb{R}^{T_l \times F}$ and the target domain $Z_l^\mathcal{T} \in \mathbb{R}^{T_l \times F}$:
\begin{equation}
    \small
    \begin{aligned}
    \mathcal{L}_{\text{DANN}}^l = &\mathcal{L}_{\text{BCE}}(D_l(Z_l^\mathcal{S}), d_\mathcal{S}) + \mathcal{L}_{\text{BCE}}(D_l(Z_l^\mathcal{T}), d_\mathcal{T}) \\
    \end{aligned}.
    \label{eq:appendix_adapted_dann1}
\end{equation}
With this, we define the final loss for the multi-resolution DANN:
\begin{equation}
    \small
    \mathcal{L}_{DANN} = \sum_{l \in L} \lambda_l \; \mathcal{L}_{DANN}^l,
    \label{eq:appendix_adapted_dann2}
\end{equation}
where $\lambda_l$ is a hyper-parameter that controls the influence of each of the resolution levels on the overall adaptation loss. Importantly, this domain classifier is trained through the use of a Gradient Reverse Layer (GRL) that is placed right before the domain classifier $D$. This allows to create a min-max game where the backbone optimizes the opposite goal of $D$, which can be seen as learning to \textit{confuse} the domain discriminator.

\textbf{ADDA~\cite{tzeng2017adversarial}:}
Following a similar approach to the previously presented extension of DANN, we adapt the subsequent work ADDA\cite{tzeng2017adversarial} to our proposed architecture. In our adaptation, this mainly affects the training loss that is employed to optimize Eq. \ref{eq:appendix_adapted_dann2}. For this, in this case, we do not include the GRL module before each of the domain classifiers but rather employ a GAN loss. This splits the original optimization into two independent objectives for the generator and the discriminator. As seen in the literature, this yields the same min-max loss but with stronger gradients, which is often beneficial. Note that we train these two objectives in an alternating manner, such that at each of the iterations we back-propagate in one or the other GAN losses.

\textbf{WDGRL~\cite{shen2018wasserstein}:}
In order to adapt WDGRL to our architecture, we again follow similar principles to those presented in the adaptation of DANN. In this case, however, we substantially modify the underlying goal following the original work \cite{shen2018wasserstein}. Concretely, in this case, the goal of $D$ is not the discriminate between the Source and Target domain, but rather to estimate the Wasserstain distance~\cite{ruschendorf1985wasserstein}.  This is, we define a level-specific discriminator $D_l:\mathbb{R}^F \rightarrow \mathbb{R}$. Integrating this module into our multi-resolution architecture yields the following loss for a given level $l$:
\begin{gather}
    \small
    \begin{aligned}
    \mathcal{L}_{\text{WDGRL}}^l = &\mathcal{L}_{\text{Wesserstein}}(D_l(Z_l^\mathcal{S}), D_l(Z_l^\mathcal{T}))\\
    \end{aligned}.
    \label{eq:appendix_adapted_wdgrl1}
\end{gather}
This yields the final loss for the multi-resolution WDGRL:
\begin{gather}
    \small
    \mathcal{L}_{WDGRL} = \sum_{l \in L} \lambda_l \; \mathcal{L}_{WDGRL}^l,
    \label{eq:appendix_adapted_wdgrl2}
\end{gather}
Importantly, following \cite{shen2018wasserstein} we enforce all domain discriminators to be 1-Lipschitz. This allows us to approximate the empirical Wasserstein distance between the source and target domain as:
\begin{gather}
\small
\begin{aligned}
    \mathcal{L}_{\text{Wesserstein}}(X^\mathcal{S}, X^\mathcal{T}) = \frac{1}{|X^\mathcal{S}|} \sum_{x^\mathcal{S} \in X^\mathcal{S}} x^\mathcal{S} - \frac{1}{|X^\mathcal{T}|} \sum_{x^\mathcal{T} \in X^\mathcal{T}} x^\mathcal{T}
\end{aligned}
\end{gather}
where $X_l^\mathcal{S},X_l^\mathcal{T}\in\mathbb{R}^{T_l}$ for each of the $T_l$ anchor embeddings and resolution level $l$.

\textbf{MSTN~\cite{xie2018learning}:} In the original work from \cite{xie2018learning}, they propose to compute an online centroid estimate of each of the class embeddings. This keeps important resemblances with our approach as they attempt to reduce the discrepancy between embeddings across domains that belong to a given action class. For this, they also require classifying each of the anchor embeddings to their corresponding class labels. In the context of UDA, this is straightforward in the case of the source domain as we have their corresponding labels. Doing so on the target domain is more challenging as this domain is unlabelled. Consequently, we follow the same strategy presented in Eq. \ref{eq:appendix_adapted_dann1} and Eq. \ref{eq:appendix_adapted_dann2}, classifying the target domain embeddings according to their pseudo-label. Hence, classifying them according to the highest confidence predicted class, if this is above a given threshold $\alpha$. This results in the source embeddings $A_i^l$ of class $i$ and resolution level $l$, and the corresponding embeddings of the target domain $B_i^l$.

We then use these embeddings to update an exponential moving average (EMA) centroid of the given class $i$ and level $l$ as follows:
\begin{equation}
    \small
    \hat{C}_{i,l}^{\mathcal{S}} = \frac{1}{|A_i^l|} \; \sum_{z \in A_i^l} z,
    \label{eq:appendix_mstn_1}
\end{equation}

\begin{equation}
    \small
    C_{i,l}^\mathcal{S} = \Theta C_{i,l}^\mathcal{S} + (1-\Theta) \hat{C}_{i,l}^{\mathcal{S}},
    \label{eq:appendix_mstn_2}
\end{equation}
where $\Theta$ is the EMA decay factor. We define the centroids of the target domain analogously:
\begin{equation}
    \small
    \hat{C}_{i,l}^{\mathcal{T}} = \frac{1}{|B_i^l|} \; \sum_{z \in B_i^l} z
\end{equation}
\label{eq:appendix_mstn_3}
\begin{equation}
    \small
    C_{i,l}^\mathcal{T} = \Theta C_{i,l}^\mathcal{T} + (1-\Theta) \hat{C}_{i,l}^{\mathcal{T}}.
    \label{eq:appendix_mstn_4}
\end{equation}
Finally, as proposed in the original paper, we apply an MSE loss to reduce the centroid discrepancy and computed the mean loss over every level and class:
\begin{equation}
    \small
    \mathcal{L}_{cent} = \frac{1}{C} \frac{1}{|L|} \sum_{i=1}^C \; \sum_{l \in L} \; MSE(C_{i,l}^\mathcal{S}, C_{i,l}^\mathcal{T}).
\end{equation}

\textbf{FGDA~\cite{gao2021gradient}:} In the original work of \cite{gao2021gradient}, the authors argue about the equilibrium issues of existing adversarial methods that attempt to reduce the divergence between feature distributions. In this context, they propose a \textit{Feature Gradient Distribution Alignment (FGDA)} approach that attempts to align the resulting gradients across the Source and Target domain, using a simple Gradient discriminator module. In our setup, this adaptation is relatively straightforward, and implies only minimal changes with respect to our previously described DANN~\cite{ganin2015unsupervised} baseline. Concretely, in DANN we fed the intermediate anchor representations $Z_l^\mathcal{S}$ and $Z_l^\mathcal{T}$ to the domain discriminator $D_l$. In this case, we leverage instead their respective gradients $\frac{\partial L_{task}}{\partial Z_l^\mathcal{T}}$ and $\frac{\partial L_{task}}{\partial Z_l^\mathcal{T}}$. These are fed as input to a gradient domain discriminator $D_l$ that has the same architecture as that used in DANN, which aims to discern the domain that produced these gradients. This results in the following loss:

\begin{equation}
    \small
    \begin{aligned}
    \mathcal{L}_{\text{FGDA}}^l = &\mathcal{L}_{\text{BCE}}(D_l(\frac{\partial L_{task}}{\partial Z_l^\mathcal{S}}), d_\mathcal{S}) + \mathcal{L}_{\text{BCE}}(D_l(\frac{\partial L_{task}}{\partial Z_l^\mathcal{T}}), d_\mathcal{T}). \\
    \end{aligned}.
    \label{eq:appendix_adapted_fgda}
\end{equation}

Note that this loss is trained in a standard adversarial fashion by incorporating a GRL module that inverts the gradients of the discriminator.

\textbf{DRDA~\cite{huang2023discriminative}:} This work observes that the features of different class labels often expand outwards in multiple directions, forming an overall radial pattern. They argue that transferring this discriminative pattern across domains can be beneficial for performance. To adapt this method to our proposed architecture, we follow a similar approach to that described in the original work and implement their \textit{Global Transformation} and a \textit{Local Refinement}. More specifically, the \textit{Global Transformation} performs a similar process to that described for the MSTN\cite{xie2018learning}, but with more global, level-wise, centroids. Hence, we compute the set of global centroids $\{C_l\}_{l\in L}^\mathcal{S}$ and $\{C_l\}_{l\in L}^\mathcal{T}$, which are defined as the mean feature embedding for each of the given resolution levels. Then, in order to reduce the expected mean feature shift, we use an MSE loss to reduce to distance.
\begin{equation}
    \small
    \begin{aligned}
    \mathcal{L}_{glob\_transf} = \frac{1}{|L|} \sum_{l\in L} MSE(C_l^\mathcal{S}, C_l^\mathcal{T})
    \end{aligned}.
\end{equation}

For the \textit{Local Refinement}, we essentially follow the same formulation described in detail in the original work \cite{huang2023discriminative}. The only adaptation that we make is that the original work operates on a single-resolution level. Thus, for our baseline we define the class-level wise centroids as described in Eq. \ref{eq:appendix_mstn_1} - \ref{eq:appendix_mstn_4}. Then, we follow exactly their formulation to define the radial structure loss described in \cite{huang2023discriminative} using our computed local centroids $\{C_l\}_{i \in C, l\in L}^\mathcal{S}$ and $\{C_l\}_{i \in C,l\in L}^\mathcal{T}$. Importantly, we define the local refinement loss for each of the levels independently and then average across levels to compute the final refinement loss.

\textbf{SSTDA~\cite{chen2020action}:} In the original work, SSTDA~\cite{chen2020action} proposes to couple what they called \textit{Local SSTDA} with \textit{Global SSTDA}. \textit{Local SSTDA} refers to the standard frame-level alignment loss. To adapt this to our multi-resolution setup, we follow the same strategy as in our adaptation of DANN, applying adversarially training a classifier that learns to discriminate the domain that each of the frames/embeddings comes from. \textit{Global SSTDA}, in contrast, refers to a sequential domain prediction loss. This splits the original source and target domain into 2 clips of sequential frames. They then apply a \textit{domain attentive temporal pooling}~\cite{chen2020action} to first weight the frames according to their entropy, and then pool them into a single compact clip embedding. The source and target clip embeddings are then randomly shuffled, and finally passed through a domain classifier that learns to predict the permutation that these clips form (see the original work \cite{chen2020action} for more details). To adapt this methodology to our multi-resolution setup, we apply this idea for every level independently. Concretely, we split into two segments the domain-invariant feature representations of domains $\mathcal{S}$ and $\mathcal{T}$ for a given level $l$, yielding $Z_l^\mathcal{S} = \{Z_{l,1}^\mathcal{S}, Z_{l,2}^\mathcal{S}\}$ and $Z_l^\mathcal{T} = \{Z_{l,1}^\mathcal{T}, Z_{l,2}^\mathcal{T} \}$. Then, following the original proposal, we apply \textit{domain attentive temporal pooling} to compute segment-level representations $V_l^\mathcal{S} = \{V_{l,1}^\mathcal{S}, V_{l,2}^\mathcal{S}\}$ and $V_l^\mathcal{T} = \{V_{l,1}^\mathcal{T}, V_{l,2}^\mathcal{T} \}$. Finally, we apply the sequential domain prediction on these segment embeddings of a given resolution level $l$. For this, we shuffle them randomly into -- e.g., $\{V_{l,1}^\mathcal{S}, V_{l,2}^\mathcal{T}, V_{l,2}^\mathcal{S}, V_{l,1}^\mathcal{T}\}$ -- and train a level-wise domain classifier $D_L:\mathbb{R}^{4F} \rightarrow \{0,1\}^{\#perm}$ to predict which of the possible permutations was used.

\textbf{TranSVAE~\cite{wei2024unsupervised}}: This work is one of the latest state-of-the-art methods on domain adaptation for video action recognition. This is especially interesting given the innovative disentanglement-based approach. This approach, concretely, leverages the use of a Variational Autoencoder that learns two types of representations: 1) Frame (clip) level representations with dynamic information and 2) Static video level representation that includes domain-specific information. The main goal of this model is to enforce domain invariance on the frame-level representations. For this, they leverage several domain adversarial losses (see the original \cite{wei2024unsupervised} for more details). Finally, it leverages a classification head to aggregate the domain invariant frame-level representations to perform a single video-level prediction.

Adapting this work to our target task, Temporal Action Localization, is however nontrivial. Notice that we deal with sparse detection setups, so it is not enough to transform the goal of TranSVAE into a frame-level classification problem. We find this Action Segmentation to be ineffective for our chosen benchmarks. Consequently, we propose instead to substitute their original classification head for the SGP pyramid and classification/localization head that we describe in Sec. \ref{sec:method_backbone} and \ref{sec:method_classification_and_localization}.  Hence, we utilize the domain invariant representations computed by TranSVAE as input of the multi-resolution pyramid and later prediction head. This makes this model architecture adaptable to our proposed experimental setup, and comparable to \textit{SADA}.
\section{Extended results and ablations}\label{sec:appendix_ablation_other_scenarios}
In this section, we expand the ablation studies presented in Sec. \ref{sec:ablation_studies} by analyzing three additional scenarios. Concretely, we compare the performance of \textit{SADA} with the adaptations of different UDA methods, and with different variations of our loss, in S1, S2 and S4.

\subsection{Extened comparison with existing UDA methods}\label{sec:appendix_ablation_da}

In Tab. \ref{tab:ablation_baselines} of the main text we compared the performance of our proposed method to existing UDA baselines when evaluated on scenario S3. To gain further insights on this study, in Tab.~\ref{tab:ablation_baselines_supp} we extend this analysis to 3 further scenarios --i.e., scenarios S1, S2 and S4. In this regard, similarly to Sec. \ref{sec:ablation_studies}, we observe that SADA improves the results of all the considered baselines for all these scenarios. More specifically, SADA attains a performance boost of up to $6.14\%$ mAP, $2.04\%$ mAP, and $6.33\%$, on the first, second and third considered scenario, respectievely. Moreover, our method improves by $0.3\%$ mAP, $0.72\%$ mAP and $0.19\%$ mAP the performance of the second best-performing model in these scenarios, respectively. Interestingly, in most of these cases, the second best-performing method is DANN\cite{ganin2016domain} which considerably improves other more complex methods like SSTDA\cite{chen2020action}. Another interesting note is that WDGRL\cite{shen2018wasserstein} and DRDA\cite{huang2023discriminative} present a considerable variability in the performance. For instance, while WDGRL is the third best-performing method in the second scenario, its performance drops drastically in the first and third scenarios attaining only a $20.13\%$ mAP and $20.51\%$ mAP, respectively. Similarly, FGDA\cite{huang2023discriminative} nearly achieves state-of-the-art in S1 but considerably drops in S4. This contrasts with the performance of similar methods like ADDA\cite{tzeng2017adversarial} or DANN\cite{ganin2016domain} that show a much more robust performance across the different scenarios.

\begin{table}[t]
\footnotesize
\centering
\begin{tabular}{c | ccc | c}
\toprule
\multicolumn{5}{c}{\textbf{S1}} \\
\midrule
\textbf{Model} & \multicolumn{3}{c|}{\textbf{mAP \{10,30,50\}\%}} & \textbf{Avg} \\
\midrule
DANN \cite{ganin2016domain} & 30.87 & 27.45 & 18.29 & 25.97 \\
ADDA \cite{tzeng2017adversarial} & 30.55 & 27.09 & 18.17 & 25.27 \\
WDGRL \cite{shen2018wasserstein} & 24.85 & 21.87 & 13.68 & 20.13 \\
FGDA \cite{gao2021gradient} & 29.34 & 26.99 & 17.54 & 24.62 \\
DRDA \cite{huang2023discriminative} & 31.50 & 27.97 & 18.79 & 26.08 \\
MSTN \cite{xie2018learning} & 30.88 & 27.44 & 18.25 & 25.52 \\
SSTDA \cite{chen2020action} & 29.92 & 26.49 & 17.55 & 24.65 \\
TranSVAE \cite{wei2024unsupervised} & 25.88 & 21.74 & 13.12 & 20.25 \\
\midrule
\textbf{Ours (SADA)} & \textbf{31.60} & \textbf{28.22} & \textbf{18.98} & \textbf{26.27} \\
\midrule

\multicolumn{5}{c}{\textbf{S2}} \\
\toprule
Model & \multicolumn{3}{c |}{mAP \{10,30,50\}\%} & Avg\\
\midrule
DANN \cite{ganin2016domain} & 31.32 & 28.55 & 20.10 & 26.65 \\ 
ADDA \cite{tzeng2017adversarial} & 30.16 & 27.58 & 19.72 & 25.82 \\
WDGRL \cite{shen2018wasserstein} & 24.77 & 21.96 & 14.81 & 20.51 \\
FGDA \cite{gao2021gradient} & 25.20 & 22.72 & 15.05 & 20.99 \\
DRDA \cite{huang2023discriminative} & 28.78 & 26.13 & 17.58 & 24.16 \\
MSTN \cite{xie2018learning} & 30.17 & 27.32 & 18.55 & 25.34  \\
SSTDA \cite{chen2020action} & 29.58 & 26.81 & 18.36 & 24.92  \\
TranSVAE \cite{wei2024unsupervised} &28.31 & 25.67 & 16.72  & 23.57  \\
\midrule
\textbf{Ours (SADA)} & \textbf{31.54} & \textbf{28.77} & \textbf{20.22} & \textbf{26.84}  \\

\midrule

\multicolumn{5}{c}{\textbf{S4}} \\
\toprule
Model & \multicolumn{3}{c |}{mAP \{10,30,50\}\%} & Avg\\
\midrule
DANN \cite{ganin2016domain} & 34.11 & 30.48 & 20.75 & 28.44\\ 
ADDA \cite{tzeng2017adversarial} & 33.51 & 29.82 & 20.05 & 27.79\\
WDGRL \cite{shen2018wasserstein} & 33.47 & 29.75 & 20.67 & 27.96\\
FGDA \cite{gao2021gradient} & 31.28 & 27.55 & 18.12 & 25.65 \\
DRDA \cite{huang2023discriminative} & 32.69 & 28.93 & 19.70 & 27.01 \\
MSTN \cite{xie2018learning} & 32.95 & 29.22 & 19.20 & 27.12 \\
SSTDA \cite{chen2020action} & 33.17 & 29.38 & 18.92 & 27.15\\
TranSVAE \cite{wei2024unsupervised} & 32.9 & 28.67 & 17.44 & 26.34 \\
\midrule
\textbf{Ours (SADA)} & \textbf{34.86} & \textbf{31.16} & \textbf{21.46} & \textbf{29.16} \\

\bottomrule
\end{tabular}
\vspace{0.3cm}
\caption{Ablation study on S1, S2 and S4, comparing the performance of our proposal \textit{SADA} with the chosen domain adaptation methods coupled in our proposed architecture.}
\vspace{-0.5cm}
\label{tab:ablation_baselines_supp}
\end{table}

\subsection{Extended ablation over variants of the SADA loss}\label{sec:appendix_ablation_variants_sada}

\begin{table}[t]
\footnotesize
\centering
\begin{tabular}{ccc|ccc|c}
\toprule
\multicolumn{7}{c}{\textbf{S1}} \\
\midrule
\textbf{Local} & \textbf{Global} & \textbf{Bkg} & \multicolumn{3}{c|}{\textbf{mAP \{10,30,50\}\%}} & \textbf{Avg} \\
\midrule
 & & & 30.74 & 27.23 & 18.07 & 25.34 \\
 & $\boldcheckmark$ & & 30.87 & 27.45 & 18.29 & 25.97  \\
 &  & $\boldcheckmark$ & 30.97 & 27.64 & 18.39 & 25.67 \\
 & $\boldcheckmark$ & $\boldcheckmark$ & 30.36 & 26.82 & 18.01 & 25.06 \\
$\boldcheckmark$ & & & 30.82 & 27.57 & 18.30 & 25.56 \\
$\boldcheckmark$ & $\boldcheckmark$ & & 30.30 & 26.98 & 17.85 & 25.04 \\
\midrule
$\boldcheckmark$ & & $\boldcheckmark$ & \textbf{31.60} & \textbf{28.22} & \textbf{18.98} & \textbf{26.27} \\

\midrule 

\multicolumn{7}{c}{\textbf{S2}}\\
\midrule
Local & Global & Bkg & \multicolumn{3}{c |}{mAP \{10,30,50\}\%} & Avg\\
\midrule
 & & & 29.65 & 26.86 & 19.14 & 25.21  \\
 & $\boldcheckmark$ & & 31.32 & 28.55 & 20.10 & 26.65 \\
 & & $\boldcheckmark$ & 29.94 & 27.39 & 19.40 & 25.57 \\
  & $\boldcheckmark$ & $\boldcheckmark$ & 30.86 & 28.02 & 19.84 & 26.57 \\
$\boldcheckmark$ & & & 30.84 & 27.98 & 19.37 & 26.06 \\
$\boldcheckmark$ & $\boldcheckmark$ & & 29.77 & 27.10 & 18.70 & 25.19\\
\midrule
$\boldcheckmark$ & & $\boldcheckmark$ & \textbf{31.54} & \textbf{28.77} & \textbf{20.22} & \textbf{26.84}  \\

\midrule
\multicolumn{7}{c}{\textbf{S4}} \\
\midrule
Local & Global & Bkg & \multicolumn{3}{c |}{mAP \{10,30,50\}\%} & Avg\\
\midrule
 & & & 34.41 & 30.58 & 21.00 & 28.66  \\
 & $\boldcheckmark$ & & 34.11 & 30.48 & 20.75 & 28.44 \\
 & & $\boldcheckmark$ & 34.46 & 30.85 & 20.90 & 28.73 \\
  & $\boldcheckmark$ & $\boldcheckmark$ & 34.63 & 31.08 & 21.26 & 28.99 \\
$\boldcheckmark$ & & & 33.47 & 29.91 & 20.17 & 27.85 \\
$\boldcheckmark$ & $\boldcheckmark$ & & 32.84 & 29.06 & 19.41 & 27.10 \\
\midrule
$\boldcheckmark$ & & $\boldcheckmark$ & \textbf{34.86} & \textbf{31.16} & \textbf{21.46} & \textbf{29.16} \\

\bottomrule
\end{tabular}
\caption{Ablation study of the effect of several variations of our proposed \textit{SADA} loss on S1, S2 and S4.}
\label{tab:ablation_loss_supp}
\end{table}

Closely related to Sec. \ref{sec:ablation_studies} of the main text, in Tab. \ref{tab:ablation_loss_supp} we extend the ablation over different variants of our loss on scenarios S1, S2 and S4. Observe in this table that our main insights from Sec. \ref{sec:ablation_studies} of the main text are mostly confirmed. Concretely, one of our main observations was that aligning the complete \textit{global} distribution was generally preferable over aligning the partial \textit{background} distribution only. This observation is satisfied in most of the scenarios that we study --with the exception of scenario S2. Similarly, in Sec. 4.3 of the main text we argued that combining alignment loss that presented overlaps yielded also a decrease in performance. More in detail, this happens when we couple either the \textit{local} or the \textit{background} loss with a \textit{global} alignment approach. In these cases, the embeddings have a double (simultaneous) alignment objective, which we argue is not desirable. As observed in Tab. \ref{tab:ablation_loss_supp}, this observation is also mostly satisfied in these scenarios as the combination of these losses yields a performance degradation in 7 of the 8 cases studied in this paper.

\section{Ablation of the effect of the $\lambda$ hyper parameters}\label{sec:appendix_effect_of_lambda_hyperparam}
As described in Eq. \ref{eq:sada_loss}, we define our SADA loss using a set of hyperparameters $\{\lambda_l\}_{l\in L}$, where each of the parameters $\lambda_l \in (0,1)$ controls the influence of a given resolution level in the overall loss. Given the relevance of these parameters, below we ablate over different variations to showcase the effect that they have in the final performance of the model. Concretely, we define the three following scenarios:
\begin{enumerate}[nosep]
    \item \textbf{All levs 1}: This scenario sets all the parameters $\lambda_l = 1$, keeping the contribution of each of the levels the same.
    \item \textbf{First 3 levs 1}: This scenario sets the first half of the levels to 1 and the rest to 0. This is $\lambda_0 = \lambda_1 = \lambda_2 = 1$ and $\lambda_3 = \lambda_4 = \lambda_5 = 0$.
    \item \textbf{Last 3 levs 1}: This scenario sets the first half of the levels to 0 and the rest to 1. This is $\lambda_0 = \lambda_1 = \lambda_2 = 0$ and $\lambda_3 = \lambda_4 = \lambda_5 = 1$.
\end{enumerate}

All in all, this study attempts to clarify how sensitive is our model to this design decision. In this regard, these results (see Tab. \ref{tab:ablation_lambda}) indicate a mild sensitivity to this choice. Specifically, we observe that the second best-performing method attains an absolute difference over the optimal choice of only $-0.08\%$ mAP and $-0.57\%$ mAP for the two considered scenarios S1 and S3, respectively. Moreover, we observe that the worst performing naive strategy still outperforms in both scenarios 4 out of the 6 tested UDA baselines (see Tab. \ref{tab:ablation_baselines}).

\begin{table}[t]
\footnotesize
\centering
\begin{tabular}{c | ccc | c  }
\toprule

\multicolumn{5}{c}{\textbf{S1}} \\
\midrule
Strategy & \multicolumn{3}{c |}{mAP \{10,30,50\}\%} & Avg \\
\midrule
All levs 1 & 30.60 & 27.36 & 18.41 & 25.45 \\
First 3 levs 1 & 31.05 & 27.64 & 18.30 & 25.66  \\
Last 3 levs 1 & 30.95 & 27.59 & 18.56 & 25.70 \\
\midrule
"Optimal choice" & \textbf{31.60} & \textbf{28.22} & \textbf{18.98} & \textbf{26.27} \\

\midrule

\multicolumn{5}{c}{\textbf{S3}} \\
\midrule
Strategy & \multicolumn{3}{c |}{mAP \{10,30,50\}\%} & Avg \\
\midrule
All levs 1 & 32.42 & 29.59 & 19.34 & 27.11  \\
First 3 levs 1 & 30.96 & 27.48 & 18.29 & 25.58 \\
Last 3 levs 1 & 30.54 & 27.19 & 18.28 & 25.34 \\
\midrule
"Optimal choice" & \textbf{32.69} & \textbf{29.17} & \textbf{19.72} & \textbf{27.19}\\

\bottomrule
\end{tabular}
\caption{Ablation of the effect of the $\lambda$ parameters, which regulate the influence of each of the resolution levels on the overall SADA loss. The "optimal choice" refers to the hyperparameter choice resulting from a Bayesian optimization process that leverages the labeled source domain to identify the best-performing set of parameters. Moreover, for efficiency purposes, we limit the search space to a uniform sampling in the (0,1] interval --i.e., $\{0.1, 0.2, \dots, 0.9, 1.0\}$.}
\vspace{-0.3cm}
\label{tab:ablation_lambda}
\end{table}
\section{Study of the class embedding}\label{sec:appendix_study_of_the_class_embedding}

\begin{table}[t]
\footnotesize
\centering
\begin{tabular}{c | ccccc | c}
\toprule

\multicolumn{7}{c}{\textbf{S1}} \\
\midrule
Strategy & \multicolumn{5}{c |}{mAP \{10,20,30,40,50\}\%} & Avg \\
\midrule
One-hot & 31.31 & 29.97 & 27.84 & 24.28 & 18.81 & 26.44\\ 
Random emb & 31.38 & 30.01 & 27.94 & 24.13 & 18.70 & 26.43 \\
Sinusoidal~\cite{vaswani2017attention} & 31.10 & 29.78 & 27.71 & 23.99 & 18.56 & 26.23 \\
\midrule
Learnable & \textbf{31.68} & \textbf{30.32} & \textbf{28.37} & \textbf{24.55} & \textbf{19.09} & \textbf{26.80} \\

\midrule

\multicolumn{7}{c}{\textbf{S3}} \\
\midrule
Strategy & \multicolumn{5}{c |}{mAP \{10,20,30,40,50\}\%} & Avg\\
\midrule
One-hot & 32.29 & 31.09 & 28.58 & 24.88 & 19.56 & 27.28 \\ 
Random emb & 30.71 & 29.50 & 27.43 & 23.78 & 18.52 & 25.99 \\
Sinusoidal~\cite{vaswani2017attention} & 29.93 & 28.57 & 26.44 & 22.87 & 17.01 & 24.96 \\
\midrule
Learnable & \textbf{32.69} & \textbf{31.49} & \textbf{29.17} & \textbf{25.51} & \textbf{19.72} & \textbf{27.72} \\
\bottomrule
\end{tabular}
\caption{Ablation of the effect of the use of a learnable class embedding over other static strategies.}
\label{tab:ablation_class_embeddings}
\end{table}

One of the main contributions of our work is to adversarially align distributions in a class-wise fashion. Intuitively, this requires that our level-wise domain discriminator \textit{knows} the class distribution that it is aligning. In this regard, in Sec. \ref{sec:method_our_proposal} we propose to concatenate to every anchor a learnable embeddings $e_i \in \mathbb{R}^{F}$ of its corresponding class $i$ (see Eq. \ref{eq:local_alignment}-\ref{eq:bkg_loss}). In this ablation, we empirically justify our choice. For this, we compare our approach against several non-learnable alternatives to encode a given class $i$: a naive one-hot encoding, a random class-wise dense  initialization, and the sinusoidal encoding originally proposed by~\cite{vaswani2017attention}.

In Tab.~\ref{tab:ablation_class_embeddings} we show the experimental results of each of the variants tested in S1 and S3. These indicate that a naive one-hot encoding of a class is the best-performing non-learnable strategy, obtaining considerable improvements over the other two tested non-learnable baselines. This improvement is especially prominent in the first scenario, boosting the performance over the other non-learnable strategies by $1.05\%$ mAP and $2.32\%$ mAP, respectively. Moreover, to our surprise, the popular Sinusoidal encoding~\cite{vaswani2017attention} proves to be the worst-performing method, being consistently outperformed by even the random dense encoding. Finally, we highlight that in both scenarios the use of a learnable class embedding yields the best results, improving by $0.44\%$ mAP and $0.36\%$ mAP the performance of the one-hot encoding strategy, which justifies our choice.

\section{Ablation per class}\label{sec:appendix_per_class}

\begin{table*}[t]
\footnotesize
\centering
\begin{tabular}{c|cccccccccc}
\hline
& \textit{take} & \textit{put} & \textit{wash} & \textit{open} & \textit{close} & \textit{insert} & \textit{turn-on} & \textit{cut} & \textit{turn-off} & \textit{pour} \\
\hline
\multirow{1}{*}{Ours (source-only)} & 24.20 & 29.30 & 37.30 & 32.69 & \textbf{23.89} & 10.98 & 37.39 & \textbf{18.18} & 26.24 & 18.55 \\
\hline
\multirow{1}{*}{DANN~\cite{ganin2016domain}} & 24.85 & 30.51 & 37.09 & \textbf{35.08} & 22.93 & \textbf{11.68} & 35.80 & 17.23 & 24.19 & \textbf{20.35} \\
\hline
\multirow{1}{*}{Ours (SADA)} & \textbf{26.31} & \textbf{30.70} & \textbf{39.14} & 34.68 & 23.42 & 10.58 & \textbf{39.86} & 17.05 & \textbf{26.58} & 18.78 \\
\hline
\end{tabular}
\vspace{0.3cm}
\caption{Per class mAPs (in percentage) obtained by the source only variation of our model, DANN~\cite{long2015learning} and our proposal \textit{SADA}. These results correspond to S1, which defines \textit{black old} kitchens as the source and the rest as a target.}
\label{tab:black_kitchens_old}
\end{table*}

\begin{table*}[t]
\footnotesize
\centering
\begin{tabular}{c|cccccccccc}
\hline
& \textit{take} & \textit{put} & \textit{wash} & \textit{open} & \textit{close} & \textit{insert} & \textit{turn-on} & \textit{cut} & \textit{turn-off} & \textit{pour} \\
\hline
\multirow{1}{*}{Ours (source-only)} & 24.11 & 25.69 & 29.77 & 31.96 & 27.8 & 6.69 & 25.52 & 34.16 & 17.48 & \textbf{29.18} \\
\hline
\multirow{1}{*}{DANN~\cite{ganin2016domain}} & 24.78 & 24.38 & 28.75 & 32.73 & 29.32 & 5.38 & \textbf{27.52} & 36.73 & 17.70 & 29.05 \\
\hline
\multirow{1}{*}{Ours (SADA)} & \textbf{28.32} & \textbf{31.25} & \textbf{31.18} & \textbf{34.52} & \textbf{31.82} & \textbf{7.29} & 26.66 & \textbf{39.12} & \textbf{18.68} & 28.34 \\
\hline
\end{tabular}
\vspace{0.3cm}
\caption{Per class mAPs (in percentage) obtained by the source only variation of our model, DANN~\cite{long2015learning} and our proposal \textit{SADA}. These results correspond to S3, which defines \textit{black new} kitchens as the source and the rest as a target.}
\label{tab:black_kitchens_new}
\end{table*}

In this section, we complement the analysis from Sec. \ref{sec:experimental_results} with the detailed class-wise metrics. Concretely, in Tab. \ref{tab:black_kitchens_old} and Tab. \ref{tab:black_kitchens_new} we show the respective class-wise mAP scores of the 10 considered classes on both S1 and S3. For the analysis, we also include the results obtained with the source-only variant of our model as well as our chosen DANN\cite{ganin2015unsupervised} baseline. 

Observe that in S3 (see Tab. \ref{tab:black_kitchens_old}) our model attains the best class-wise performance in 5 of the classes, while DANN~\cite{ganin2015unsupervised} does so on 3, and the source-only model on 2. In contrast, in S1 (see Tab.~\ref{tab:black_kitchens_new}) our method obtains a much clearer improvement over the chosen baselines, yielding the best results in 8 of the 10 classes. Overall, we can observe that our method performs very well in the 3 majority classes of both scenarios, where we highlight the absolute improvement of SADA over DANN~\cite{ganin2015unsupervised} of $6.8\%$ mAP for the class \textit{put} in S3. Moreover, our method fails to improve the performance of action \textit{pour} in both scenarios, which we attribute to the lack of sufficient data to operate on our proposed methodology. We also highlight that as observed in Sec. \ref{sec:ablation_studies}, S1 presents a more challenging setup, degrading the performance in other actions such as \textit{open}, \textit{close}, \textit{insert} or \textit{cut}. This might indicate the existence of intrinsic qualitative aspects that harden the adaptation when dealing with \textit{old} videos. This is not the case in Scenario 3, as aside from the aforementioned \textit{pour} segments, it only fails to achieve the best results for the \textit{turn-on} action.

\section{Ablation of the effect of \textit{background anchors}}\label{sec:appendix_ablation_negative_anchors}

In this work we focus on applying domain adaptation over anchor-based localization methods, the current state-of-the-art architectures on datasets like EpicKitchens100~\cite{Damen2018EPICKITCHENS}. The success of these architectures is normally explained by the generation of a very considerable number of anchors, largely surpassing the number of actual ground-truth labels to predict. As described in Sec. \ref{sec:method}, these are the excess anchors --i.e., \textit{background anchors}-- that are not directly matched with a GT segment, consequently training the model to predict class 0 instead. In other words, in these cases we hope for the model to learn to identify these anchor embeddings as \textit{background}, not predicting any action class during inference. This, however, is a cumbersome issue that most of the anchor-based methods face, and which results in a challenging model transfer to a new data domain. It is not clear, however, weather the use domain adaptation methods effectively help to mitigate the confusion that these embeddings induce, which we largely attribute to a high intra-class variance and a low inter-class variance with the other action classes. Hence, even though we leave a more thorough analysis as future work, in this ablation we aim to provide minimal proof of how our method partially mitigates this issue, and also of the potential that these methods would have could we manage to solve this issue.

\subsection{Masking out the anchors during inference}

{For this, in this section we firstly extend to scenario S1 the experiment presented in Sec. \ref{sec:ablation_studies} of the main text (see Tab. \ref{tab:ablation_effect_background_anchors}), where we propose an experiment where mask out the set of all the \textit{background anchors} during inference. Put differently, we leave all the training routines unchanged -- thus considering all the anchors-- while not considering the \textit{background anchors} at all during inference. This measures the predicting potential of a model that is able to perfectly deal with these embeddings. Observe in Tab. \ref{tab:extended_ablation_background_embs_s1} that in this scenario, our proposed method SADA has a potential increase of $11.05\%$ mAP if we mask all the \textit{background anchors} during inference. This contrasts with the $13.05\%$ mAP increase in the source-only version, which shows that SADA reduces this gap by $2\%$ mAP. In other words, SADA partially improves the transferability of these anchors across domains.

\begin{table}[t]
\centering
\footnotesize
    \renewcommand{\arraystretch}{0.9}
\setlength{\tabcolsep}{4pt} 
\resizebox{0.45\textwidth}{!}{%
\begin{tabular}{c|c|ccc|c|c}
\toprule
\textbf{Method} & \textbf{Mask \textit{bkg anchors}} & \multicolumn{3}{c|}{\textbf{mAP \{10,30,50\}\%}} & \textbf{Avg} & \textbf{Perf. gap} \\
\midrule
\multirow{2}{*}{Ours(src-only)} & & 30.74 & 27.23 & 18.07 & 25.38 & - \\
& \boldcheckmark & \textbf{42.34} & \textbf{42.06} & \textbf{30.91}  & \textbf{38.43} & 13.05  \\
\midrule
\multirow{2}{*}{Ours(SADA)} & & 31.60 & 28.22 & 18.98 & 26.26 & - \\
 & \boldcheckmark & 41.68 & 40.23 & 30.03 & 37.31 & \textbf{11.05} \\
 \bottomrule
\end{tabular}}%
\vspace{-0.1cm}
\caption{Ablation of the effect of the \textit{background anchors} in the performance of the model on S1.}
\vspace{-0.4cm}
\label{tab:extended_ablation_background_embs_s1}
\end{table}

\begin{table}
\footnotesize
\centering
\begin{tabular}{c | ccccc | c}
\toprule

\multicolumn{7}{c}{\textbf{S1}} \\
\midrule
Filtered \% & \multicolumn{5}{c |}{mAP \{10,20,30,40,50\}\%} & Avg \% \\
\midrule
0\% & 31.60 & 30.29 & 28.22 & 24.47 & 18.98 & 26.71 \\ 
25\% & 32.76 & 31.60 & 29.81 & 26.14 & 20.24 & 28.11 \\ 
50\% & 35.25 & 34.25 & 32.68 & 29.25 & 23.01 & 30.89 \\ 
75\% & 38.39 & 37.77 & 36.40 & 33.19 & 26.33 & 34.42 \\ 
100\% & \textbf{41.68} & \textbf{40.59} & \textbf{40.23} & \textbf{34.26} & \textbf{30.03} & \textbf{37.35} \\ 

\midrule

\multicolumn{7}{c}{\textbf{S3}} \\
\midrule
Masked \% & \multicolumn{5}{c |}{mAP \{10,20,30,40,50\}\%} & Avg \% \\
\midrule
0\% & 32.69 & 31.49 & 29.17 & 25.51 & 19.72 & 27.72 \\ 
25\% & 32.54 & 31.87 & 30.06 & 26.99 & 21.91 & 28.67 \\ 
50\% & 34.46 & 33.55 & 31.29 & 27.59 & 21.18 & 29.61 \\ 
75\% & 35.09 & 34.53 & 32.97 & 30.05 & 23.97 & 31.32 \\ 
100\% & \textbf{35.70} & \textbf{35.24} & \textbf{34.09} & \textbf{30.72} & \textbf{24.15} & \textbf{31.98} \\ 

\bottomrule
\end{tabular}
\caption{Ablation of the effect of masking out different percentages of \textit{background anchors} during \underline{inference} only.}
\label{tab:ablation_no_class1}
\end{table}

Nevertheless, we highlight that in this scenario in particular, unlike in S3, there is still a big gap which indicates that there is considerable room for improvement. To gain further insights on this issue, we additionally propose an experiment where we progressively mask out the set of \textit{background anchors}. For this, in Tab. \ref{tab:ablation_no_class1} we randomly mask out these ensuring that only a percentage of the \textit{background anchors} are not considered during inference. As observed in Tab. \ref{tab:ablation_no_class1}, in both considered scenarios we observe a linear performance increase, reaching an absolute improvement of $4.26\%$ mAP and $10.64\%$ mAP of the version with $100\%$ filtered background anchors over the $0\%$ version. This is, the variant where only the \textit{action} anchors are considered over that containing both \textit{action} and all the \textit{background} ones.

\subsection{Masking out the anchors during training and inference}

In Tab. \ref{tab:ablation_no_class2} we propose an alternative analysis. In this case, we do not only mask the \textit{background anchors} during inference but also during training. In this case, for instance, masking out $50\%$ of the \textit{background anchors} additionally implies not training any of the task losses on them, nor using them on our alignment loss. As observed in the results, masking them out entirely -- i.e., $100\%$ -- has an even bigger impact, reaching an absolute improvement of $6.1\%$ mAP and $16.52\%$ mAP, respectively, when comparing it to the $0\%$ filtering version. Nevertheless, in this case, we observe that the improvement is no longer linear, and even worsens the performance for small percentages of masking. We hypothesize this is induced by a reduction of the training data of \textit{background class}. Thus, while masking some \textit{background} embeddings has a positive effect, we must also consider the negative effect that having less training data has on the ones that we do consider. Hence, the improvement in performance remains little to non-existent until this trade-off is dominated by the positive effect of masking out the majority of \textit{background anchors}. In short, these additional experiments emphasize the need for devising carefully designed models that can properly cope with these intrinsic limitations of anchor-based methods.

\begin{table}
\footnotesize
\centering
\begin{tabular}{c | ccccc | c}
\toprule

\multicolumn{7}{c}{\textbf{S1}} \\
\midrule
Filtered \% & \multicolumn{5}{c |}{mAP \{10,20,30,40,50\}\%} & Avg \\
\midrule
0\% & 31.60 & 30.29 & 28.22 & 24.47 & 18.98 & 26.71 \\ 
25\% & 32.25 & 30.46 & 28.04 & 24.54 & 19.03 & 26.86 \\ 
50\% & 33.43 & 32.06 & 29.98 & 26.44 & 20.55 & 28.49 \\ 
75\% & 38.26 & 37.33 & 35.72 & 32.53 & 25.61 & 33.89 \\ 
100\% & \textbf{47.08} & \textbf{46.69} & \textbf{45.65} & \textbf{42.31} & \textbf{34.42} & \textbf{43.23} \\ 

\midrule

\multicolumn{7}{c}{\textbf{S3}} \\
\midrule
Masked \% & \multicolumn{5}{c |}{mAP \{10,20,30,40,50\}\%} & Avg \\
\midrule
0\% & 32.69 & 31.49 & 29.17 & 25.51 & 19.72 & 27.72 \\ 
25\% & 31.48 & 30.34 & 27.97 & 24.07 & 18.93 & 26.56\\ 
50\% & 32.11 & 31.11 & 29.25 & 25.85 & 20.72 & 27.81 \\ 
75\% & 35.99 & 35.28 & 33.48 & 29.90 & 23.64 & 31.66\\ 
100\% & \textbf{37.06} & \textbf{35.73} & \textbf{32.34} & \textbf{26.45} & \textbf{33.82} & \textbf{47.08} \\ 

\bottomrule
\end{tabular}
\caption{Ablation of the effect of masking out different percentages of \textit{background anchors} during \underline{training and inference.}}
\label{tab:ablation_no_class2}
\end{table}

\section{Additional qualitative results}\label{sec:appendix_qualitative_results}
In this section we extend our qualitative analysis from Sec. \ref{sec:ablation_studies}. To this end, we first provide in Sec.~\ref{sec:appendix_segment_visualization} additional segment visualizations. Then, in Sec.~\ref{sec:appendix_extended_tsne_analysis} we extend the analysis of the TSNE plots to study the effect of multi-resolution in the overall alignment of embeddings.

\subsection{Segment visualization}\label{sec:appendix_segment_visualization}

\begin{figure}[t]
\centering
\includegraphics[width=\columnwidth]{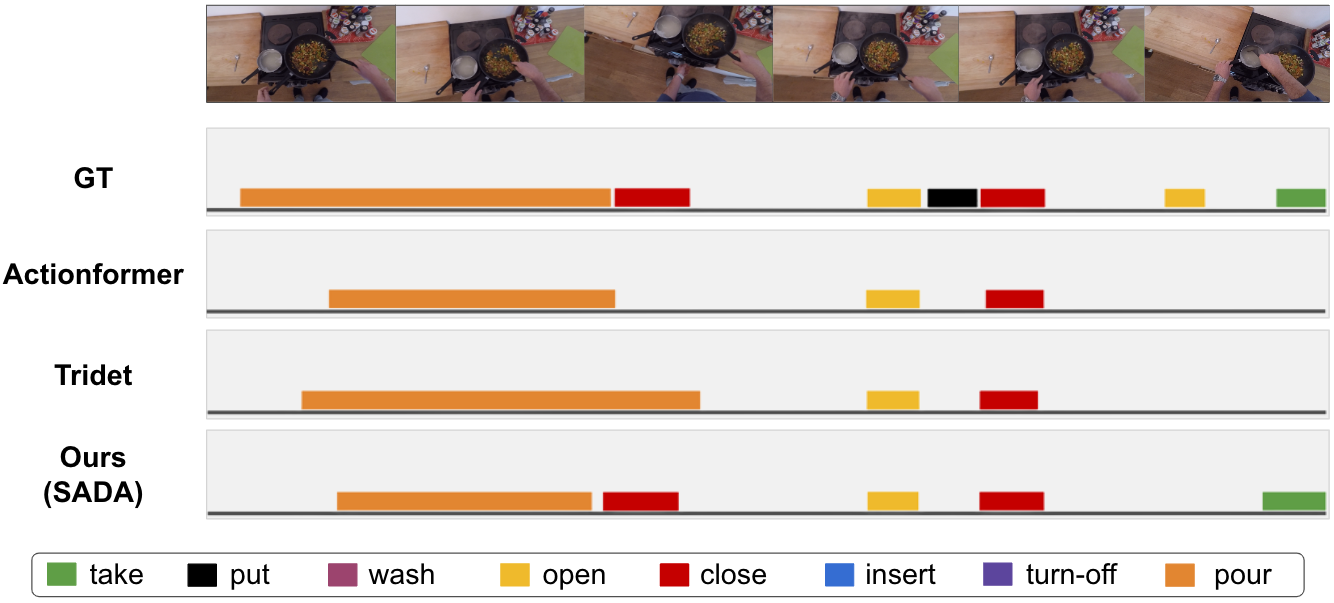}
\caption{\label{fig:overview_architecture} Visualization of the predicted segments on S1 -- i.e., using \textit{dark old kitchens} as source -- of our method and the chosen set of source-only baselines. We include on top the ground-truth (GT) segments as a reference.}
\label{fig:video_qualitative_results2}
\end{figure}

\begin{figure}[t]
\centering
\includegraphics[width=\columnwidth]{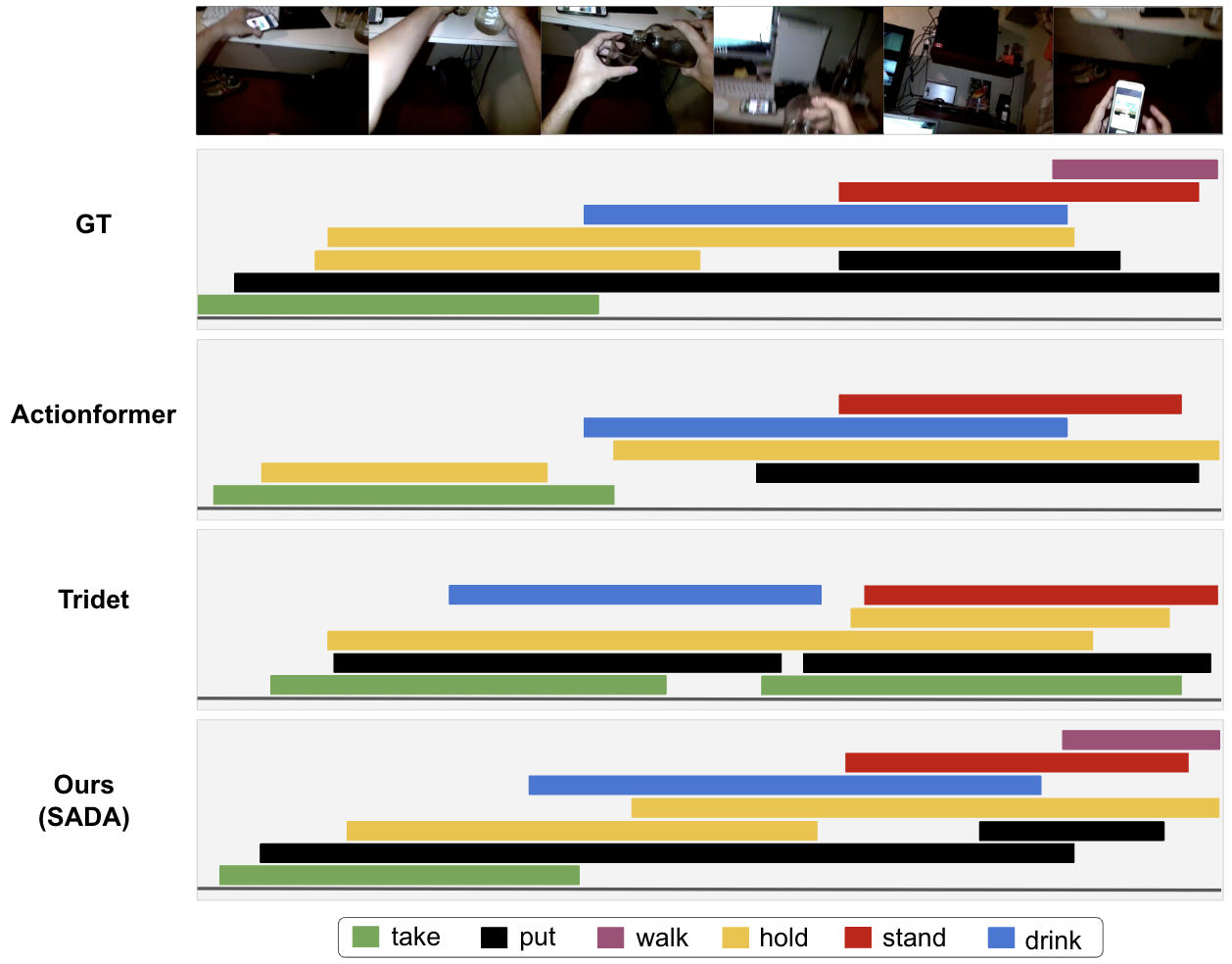}
\caption{\label{fig:overview_architecture} Visualization of the predicted segments on CharadesEgo of our method and the chosen set of source-only baselines. We include on top the ground-truth (GT) segments as a reference.}
\label{fig:video_qualitative_results3}
\end{figure}

In Fig. \ref{fig:video_qualitative_results2} we present a segment visualization of S3 -- i.e., using \textit{old} dark-counter kitchens as the source domain, and the rest as the target. Observe that in this case, both ActionFormer~\cite{zhang2022actionformer} and Tridet~\cite{shi2023tridet} perform similarly. Concretely, they mainly miss one \textit{close} action at the beginning, a \textit{put} in the middle section of the video, and finally the last two actions \textit{open} and \textit{take}. In contrast, our model is able to correctly detect the previously missing \textit{close}, and the final \textit{take}. In short, reducing by half the number of undetected actions in the video.

Similarly Fig. \ref{fig:video_qualitative_results3} shows a segment visualization that compares the chosen baselines with SADA when applied to the CharadesEgo dataset~\cite{sigurdsson2018charades}. Observe that this visualization presents a much more complex setup where there are numerous overlapping actions. In this regard, we highlight that the worst-performing model is the ActionFormer~\cite{zhang2022actionformer}. This model misses the action \textit{walk} and one of the actions \textit{put}. It also presents important limitations in terms of localization, especially for the action labels \textit{drink} and \textit{hold}. The Tridet~\cite{shi2023tridet} considerably improves this method. Nevertheless, it still misses the action \textit{walk} and is unable to locate the longest \textit{put} action. Additionally, Tridet predicts an action \textit{take} that does not correspond to any ground truth. Our method, finally, presents much more accurate scores. Importantly, our method does not miss any of the ground-truth actions and in most cases reports considerably accurate segments. We only highlight the persisting confusion in the two actions \textit{hold} which is consistent across all the tested baselines.

\subsection{Extended TSNE analysis: Studying the effect of multi resolution}\label{sec:appendix_extended_tsne_analysis}

\begin{figure}

\centering
    \begin{subfigure}[b]{0.12\textwidth}
    \centering
    \includegraphics[width=2cm]{images/tsne/src_only/tsne_lev_0_class_1.png}
    \caption{Level1(SO)\label{fig:image1}}
    \end{subfigure}
\quad
    \begin{subfigure}[b]{0.12\textwidth}
    \centering
    \includegraphics[width=2cm]{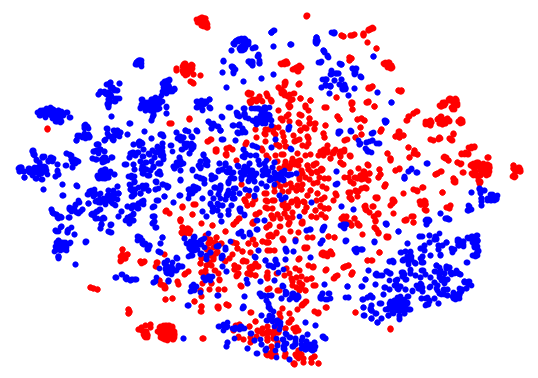}
    \caption{Level2(SO)\label{fig:image2}}
    \end{subfigure}
\quad
    \begin{subfigure}[b]{0.12\textwidth}
    \centering
    \includegraphics[width=2cm]{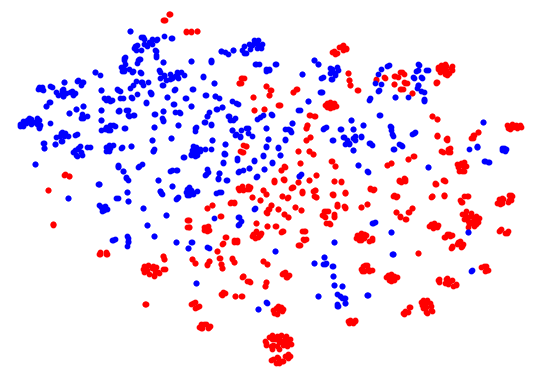}
    \caption{Level3(SO)\label{fig:image3}}
    \end{subfigure}
\hfill\\
    \begin{subfigure}[b]{0.12\textwidth}
    \centering
    \includegraphics[width=2cm]{images/tsne/SADA/tsne_lev_0_class_1.png}
    \caption{Level1(DA)\label{fig:image5}}
    \end{subfigure}
\quad
    \begin{subfigure}[b]{0.12\textwidth}
    \centering
    \includegraphics[width=2cm]{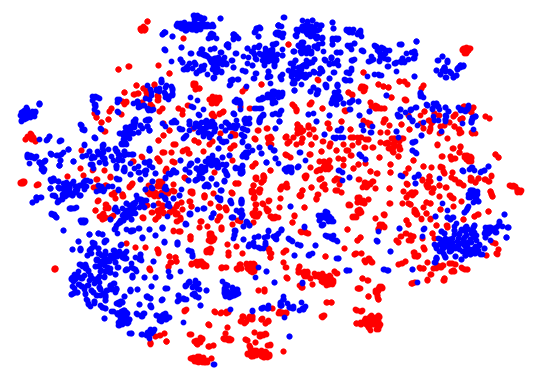}
    \caption{Level2(DA)\label{fig:image6}}
    \end{subfigure}
\quad
    \begin{subfigure}[b]{0.12\textwidth}
    \centering
    \includegraphics[width=2cm]{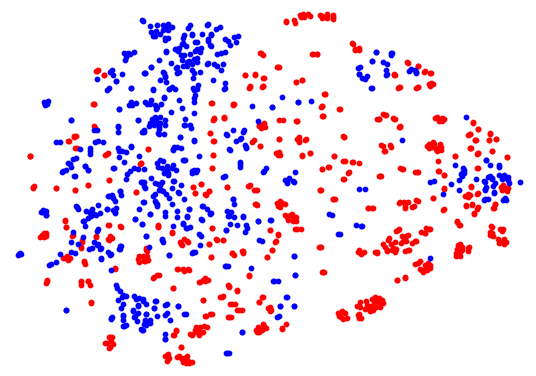}
    \caption{Level3(DA)\label{fig:image7}}
    \end{subfigure}

\caption{TSNE plots of class 1 of the source-only variation of our model (top row) and our proposed model (bottom row). Concretely, find in the 3 columns the TSNE plots of class 1 on the first 3 resolution levels of the source (red) and target (blue) domain anchors.}
\label{fig:tsne_plots_class_1}
\end{figure}

\begin{figure}

\centering
    \begin{subfigure}[b]{0.12\textwidth}
    \centering
    \includegraphics[width=2cm]{images/tsne/src_only/tsne_lev_0_class_2.png}
    \caption{Level1(SO)\label{fig:image1}}
    \end{subfigure}
\quad
    \begin{subfigure}[b]{0.12\textwidth}
    \centering
    \includegraphics[width=2cm]{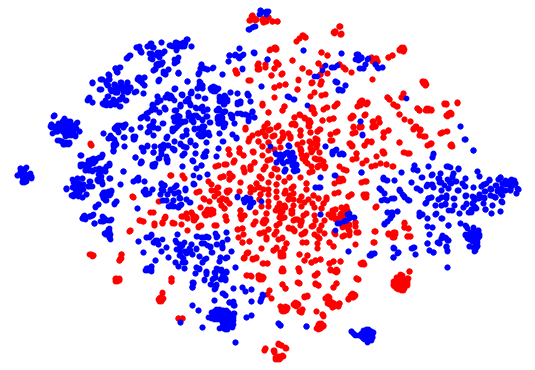}
    \caption{Level2(SO)\label{fig:image2}}
    \end{subfigure}
\quad
    \begin{subfigure}[b]{0.12\textwidth}
    \centering
    \includegraphics[width=2cm]{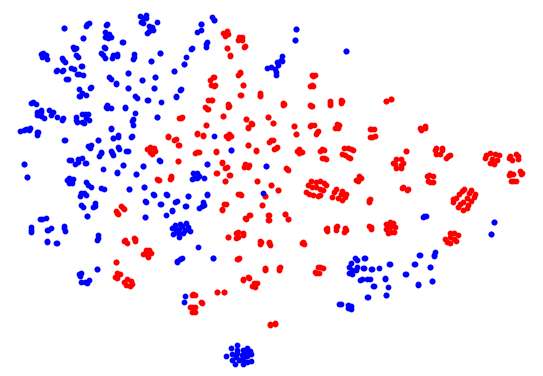}
    \caption{Level3(SO)\label{fig:image3}}
    \end{subfigure}
\hfill\\
    \begin{subfigure}[b]{0.12\textwidth}
    \centering
    \includegraphics[width=2cm]{images/tsne/SADA/tsne_lev_0_class_2.png}
    \caption{Level1(DA)\label{fig:image5}}
    \end{subfigure}
\quad
    \begin{subfigure}[b]{0.12\textwidth}
    \centering
    \includegraphics[width=2cm]{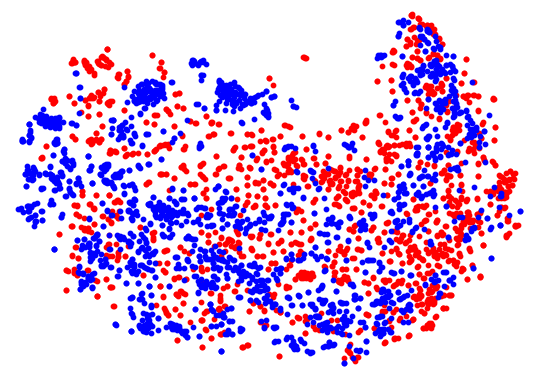}
    \caption{Level2(DA)\label{fig:image6}}
    \end{subfigure}
\quad
    \begin{subfigure}[b]{0.12\textwidth}
    \centering
    \includegraphics[width=2cm]{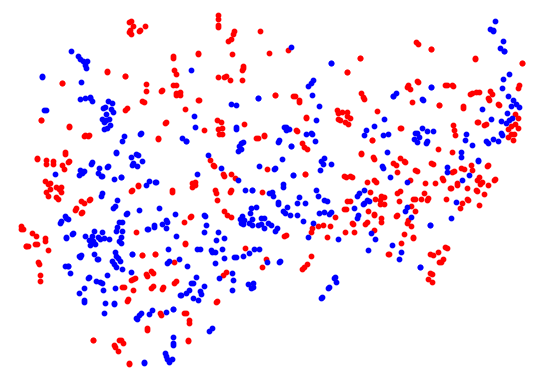}
    \caption{Level3(DA)\label{fig:image7}}
    \end{subfigure}

\caption{TSNE plots of class 2 of the source-only variation of our model (top row) and our proposed model (bottom row). Concretely, find in the 3 columns the TSNE plots of class 2 on the first 3 resolution levels of the source (red) and target (blue) domain anchors.}
\label{fig:tsne_plots_class2}
\end{figure}

\begin{figure}

\centering
    \begin{subfigure}[b]{0.12\textwidth}
    \centering
    \includegraphics[width=2cm]{images/tsne/src_only/tsne_lev_0_class_3.png}
    \caption{Level1(SO)\label{fig:image1}}
    \end{subfigure}
\quad
    \begin{subfigure}[b]{0.12\textwidth}
    \centering
    \includegraphics[width=2cm]{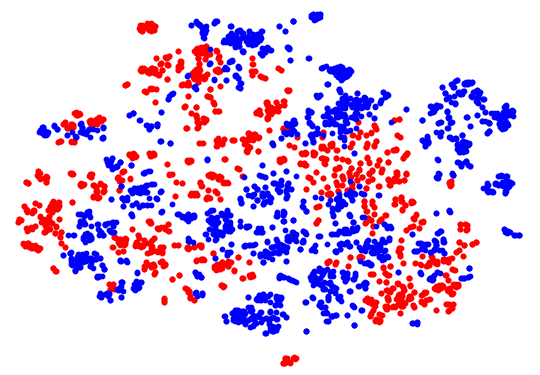}
    \caption{Level2(SO)\label{fig:image2}}
    \end{subfigure}
\quad
    \begin{subfigure}[b]{0.12\textwidth}
    \centering
    \includegraphics[width=2cm]{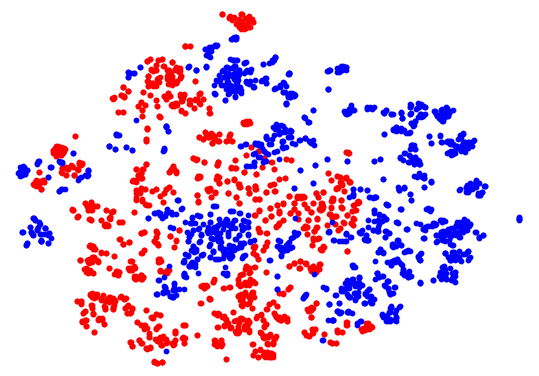}
    \caption{Level3(SO)\label{fig:image3}}
    \end{subfigure}
\hfill\\
    \begin{subfigure}[b]{0.12\textwidth}
    \centering
    \includegraphics[width=2cm]{images/tsne/SADA/tsne_lev_0_class_3.png}
    \caption{Level1(DA)\label{fig:image5}}
    \end{subfigure}
\quad
    \begin{subfigure}[b]{0.12\textwidth}
    \centering
    \includegraphics[width=2cm]{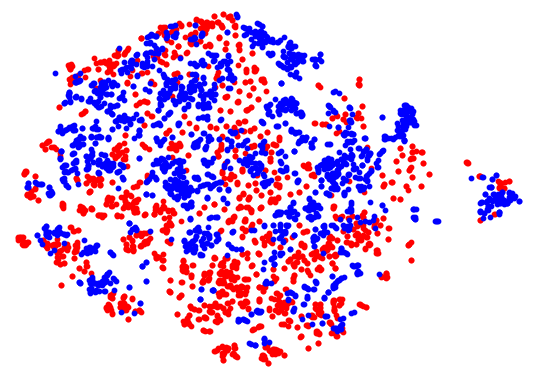}
    \caption{Level2(DA)\label{fig:image6}}
    \end{subfigure}
\quad
    \begin{subfigure}[b]{0.12\textwidth}
    \centering
    \includegraphics[width=2cm]{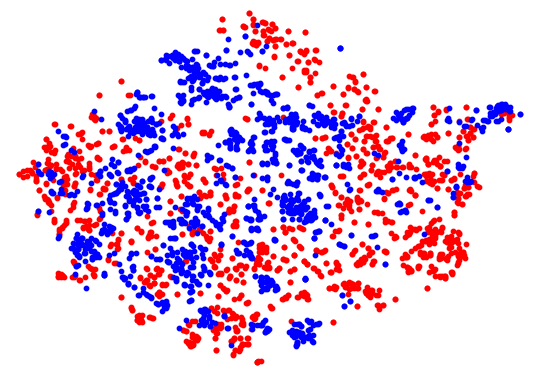}
    \caption{Level3(DA)\label{fig:image7}}
    \end{subfigure}

\caption{TSNE plots of class 3 of the source-only variation of our model (top row) and our proposed model (bottom row). Concretely, find in the 3 columns the TSNE plots of class 3 on the first 3 resolution levels of the source (red) and target (blue) domain anchors.}
\label{fig:tsne_plots_class_3}
\end{figure}

\begin{figure}

\centering
    \begin{subfigure}[b]{0.12\textwidth}
    \centering
    \includegraphics[width=2cm]{images/tsne/src_only/tsne_lev_0_no_class.png}
    \caption{Level1(SO)\label{fig:image1}}
    \end{subfigure}
\quad
    \begin{subfigure}[b]{0.12\textwidth}
    \centering
    \includegraphics[width=2cm]{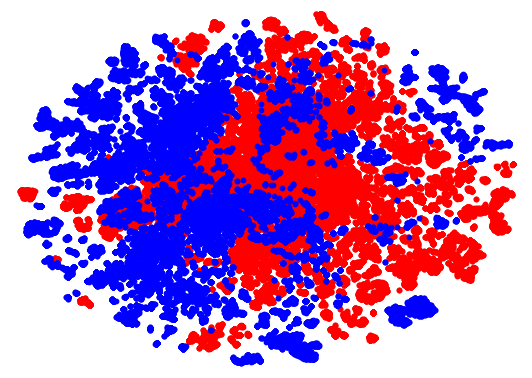}
    \caption{Level2(SO)\label{fig:image2}}
    \end{subfigure}
\quad
    \begin{subfigure}[b]{0.12\textwidth}
    \centering
    \includegraphics[width=2cm]{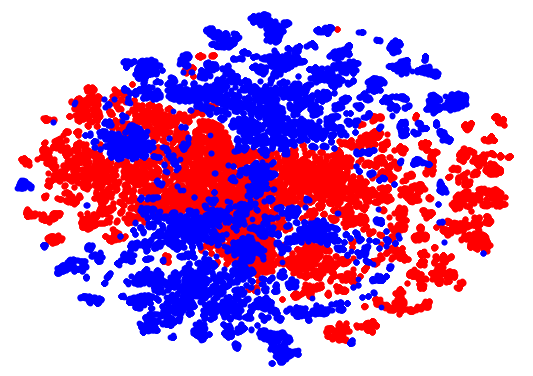}
    \caption{Level3(SO)\label{fig:image3}}
    \end{subfigure}
\hfill\\
    \begin{subfigure}[b]{0.12\textwidth}
    \centering
    \includegraphics[width=2cm]{images/tsne/SADA/tsne_lev_0_no_class.png}
    \caption{Level1(DA)\label{fig:image5}}
    \end{subfigure}
\quad
    \begin{subfigure}[b]{0.12\textwidth}
    \centering
    \includegraphics[width=2cm]{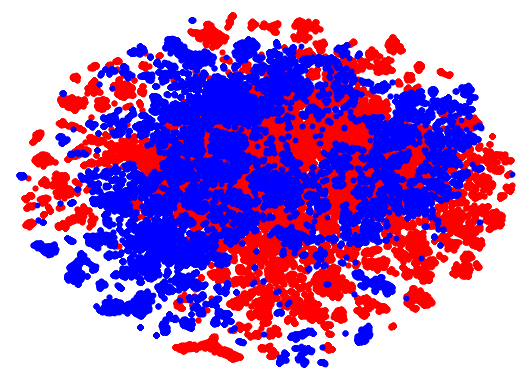}
    \caption{Level2(DA)\label{fig:image6}}
    \end{subfigure}
\quad
    \begin{subfigure}[b]{0.12\textwidth}
    \centering
    \includegraphics[width=2cm]{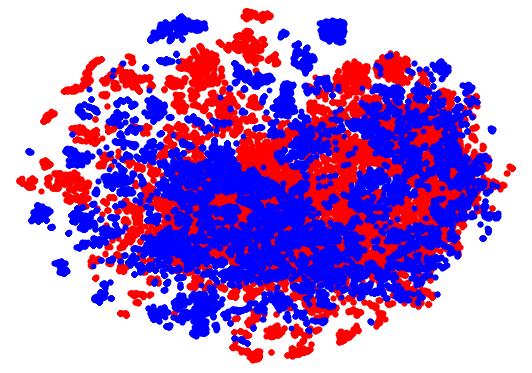}
    \caption{Level3(DA)\label{fig:image7}}
    \end{subfigure}

\caption{TSNE plots of \textit{background class} of the source-only variation of our model (top row) and our proposed model (bottom row). Concretely, find in the 3 columns the TSNE plots of \textit{background class} on the first 3 resolution levels of the source (red) and target (blue) domain anchors.}
\label{fig:tsne_plots_no_class}
\end{figure}

In the qualitative study presented in the main paper (see Fig. \ref{fig:tsne_plots}) we only include visualizations of the embeddings that correspond to the level 0 of our multi-resolution architecture. It remains unclear, however, whether this alignment behavior remains consistent across the different resolution levels. To showcase this behavior, in this section we first include in Fig.~\ref{fig:tsne_plots_class_1}~- ~Fig.~\ref{fig:tsne_plots_no_class} the TSNE plots of the 3 majority classes and of background embeddings, respectively. Each of these figures includes the different plots at each of the first 3 resolution levels, where level 0 is the shallowest, and level 3 is the deepest studied level. We highlight that deeper levels are not meaningful to plot given the scarce number of resulting domain-invariant embeddings, caused by the downsampling that each of the levels of the architecture performs.

Observe that the influence of the alignment loss follows a similar pattern in all the resolutions. Concretely, in the three presented, we observe a considerable mixing between the source and target distributions (red and blue, respectively). This contrasts with the very clear disentanglement of the representations of both domains in the source-only version, where these present little to no overlap. We highlight what seems to be the only exception which is resolution level 3 of class 1. This one shows little improvement in the mixing of the distributions compared to the source-only variant. We also find that our overall improvement of the mixing of the distributions is consistent when analyzing the \textit{background class} embeddings. Observe in Fig.~\ref{fig:tsne_plots_no_class} that in all 3 studied resolution levels, \textit{SADA} improves very considerably the alignment, pushing the feature space of both domains to be domain invariant. This emphasizes the positive influence of the background alignment term of our loss (see Eq. \ref{eq:bkg_loss}).

\end{document}